\documentclass{article}
\usepackage{microtype}
\usepackage{graphicx}
\usepackage{subfigure}
\usepackage{booktabs}

\usepackage{hyperref}

\usepackage[accepted]{icml2021}

\usepackage{amssymb}
\usepackage{amsmath}
\usepackage{enumerate}
\usepackage{multirow}
\usepackage{algorithm}
\usepackage{algorithmic}
\usepackage{tabularx}
\usepackage{xcolor,colortbl}
\usepackage{nicefrac}

\newtheorem{lemma}{Lemma}[section]
\newtheorem{proposition}{Proposition}[section]

\newenvironment{proof}{\textit{Proof.}}{\hfill$\square$}

\newtheorem{innercustomprop}{Proposition}
\newenvironment{customprop}[1]{\renewcommand\theinnercustomprop{#1}\innercustomprop}{\endinnercustomprop}
\newtheorem{innercustomlemma}{Lemma}
\newenvironment{customlemma}[1]{\renewcommand\theinnercustomlemma{#1}\innercustomlemma}{\endinnercustomlemma}

\newcommand{\inner}[1]{\left\langle#1\right\rangle}
\def\R{\mathbb{R}}
\def\N{\mathbb{N}}

\newcommand{\norm}[1]{\left\|#1\right\|}

\def\Pr{\mathrm{P}}

\def\B{\mathcal{B}}
\def\U{\mathcal{U}}
\def\P{\mathcal{P}}

\def\Pr{\mathrm{P}}
\def\Nc{\mathcal{N}}

\def\Exp{\mathbb{E}}

\def\ones{\mathop{\rm e}\nolimits}

\def\argmax{\mathop{\rm arg\,max}\limits}
\def\argmin{\mathop{\rm arg\,min}\limits}
\def\minop{\mathop{\rm min}\limits}

\def\sign{\mathop{\rm sign}\limits}

\def\min{\mathop{\rm min}\nolimits}
\def\max{\mathop{\rm max}\nolimits}
\def\ones{\mathbf{1}}

\def\niter{$N_\textrm{iter}$}

\newcolumntype{C}[1]{>{\centering\arraybackslash}p{#1}}
\newcolumntype{L}[1]{>{\raggedright\arraybackslash}p{#1}}
\newcolumntype{R}[1]{>{\raggedleft\arraybackslash}p{#1}}

\newcommand{\iter}[2]{#1^{(#2)}}

\newcommand{\hl}[1]{
		#1
	}

\def\SPSB#1#2{\rlap{\textsuperscript{\textcolor{black}{#1}}}\SB{#2}}
\def\SP#1{\textsuperscript{\textcolor{black}{#1}}}
\def\SB#1{\textsubscript{\textcolor{black}{#1}}}

\icmltitlerunning{Mind the Box: $l_1$-APGD for Sparse Adversarial Attacks on Image Classifiers}

\begin{document}

\twocolumn[
\icmltitle{Mind the Box: $l_1$-APGD for Sparse Adversarial Attacks on Image Classifiers}

\icmlsetsymbol{equal}{*}

\begin{icmlauthorlist}
\icmlauthor{Francesco Croce}{to}
\icmlauthor{Matthias Hein}{to}
\end{icmlauthorlist}

\icmlaffiliation{to}{
University of T\"{u}bingen}

\icmlcorrespondingauthor{F. Croce}{francesco.croce@uni-tuebingen.de}

\icmlkeywords{Machine Learning, ICML}

\vskip 0.3in]

\printAffiliationsAndNotice{} 

\begin{abstract}
We show that when taking into account also the image domain $[0,1]^d$, established $l_1$-projected gradient descent (PGD) attacks are suboptimal as they do not consider that the effective threat model is the intersection of the $l_1$-ball and $[0,1]^d$. We study the expected sparsity of the steepest descent step for this effective threat model and show that the exact projection onto this set is computationally feasible and yields better performance. Moreover, we propose an adaptive form of PGD which is highly effective even with a small budget of iterations. Our resulting $l_1$-APGD is a strong white-box attack showing that prior works overestimated their $l_1$-robustness. Using $l_1$-APGD for adversarial training we get a robust classifier with SOTA $l_1$-robustness. Finally, we combine $l_1$-APGD and an adaptation of the Square Attack to $l_1$ into $l_1$-AutoAttack, an ensemble of attacks which reliably assesses adversarial robustness for the threat model of $l_1$-ball intersected with $[0,1]^d$.

\end{abstract}

\section{Introduction}

The application of machine learning in safety-critical systems requires reliable decisions. Small adversarial perturbations \cite{SzeEtAl2014,KurGooBen2016a}, changing the decision of a classifier, without changing the semantic content of the image are a major problem. While adversarial training \cite{MadEtAl2018} and recent variations and improvements \cite{CarEtAl19,gowal2020uncovering,wu2021wider} are a significant progress, most proposed defenses not involving some form of adversarial training turn out to be non-robust \cite{CarWag2016,AthEtAl2018}. While the community so far has focused mainly on $l_\infty$- and $l_2$-perturbations, $l_1$-perturbation sets are complementary as they lead to very sparse changes which leave effectively most of the image unmodified and thus should also not lead to a change in the decision. While there exist a set of $l_1$-based attacks \cite{CheEtAl2018,ModEtAl19, brendel2019accurate,CroHei2019,rony2020augmented}, in contrast to the $l_\infty$- and $l_2$-case the classical white-box projected gradient descent (PGD) attack of \cite{MadEtAl2018} has not an established standard form \cite{TraBon2019,maini2020adversarial}. Moreover, training $l_1$-robust models with adversarial training has been reported to be difficult 
\cite{maini2020adversarial,liu2020towards}.

In this paper we identify reasons why the current versions of $l_1$-PGD attacks are weaker than SOTA $l_1$-attacks \cite{CheEtAl2018,CroHei2019,rony2020augmented}. A key issue is that in image classification we have the additional constraint that the input has to lie in the box $[0,1]^d$ and thus the effective threat model is the intersection of the $l_1$-ball and $[0,1]^d$. However, current $l_1$-PGD attacks only approximate the correct projection onto this set \cite{TraBon2019} and argue for a steepest descent direction without taking into account the box constraints. We first show that the correct projection onto the intersection can  be computed in  essentially the same time as the projection onto the $l_1$-ball, and then we discuss theoretically and empirically that using the approximate projection leads to a worse attack as it cannot access certain parts of the threat model. Moreover, we derive the correct steepest descent step for the intersection of $l_1$-ball and $[0,1]^d$ which motivates an adaptive sparsity of the chosen descent direction. Then, inspired by the recent work on Auto-PGD (APGD) \cite{croce2020reliable} for $l_2$ and $l_\infty$, we design a novel fully adaptive parameter-free PGD scheme so that the user does not need to do step size selection for each defense separately which is known to be error prone. Interestingly, using our $l_1$-APGD we are able to train the model with the highest $l_1$-robust accuracy for $\epsilon=12$ while standard PGD fails due to catastrophic overfitting \cite{Wong2020Fast} and/or overfitting to the sparsity of the standard PGD attack. Finally, following \cite{croce2020reliable} we assemble  $l_1$-APGD for two different losses, the targeted $l_1$-FAB attack \cite{CroHei2019}, and an $l_1$-adaptation of the SOTA black-box Square Attack \cite{ACFH2019square} into a novel parameter-free $l_1$-AutoAttack which leads to a reliable and effective assessment of $l_1$-robustness similar to AutoAttack \cite{croce2020reliable} for the $l_2$- and $l_\infty$-case. All proofs can be found in App.~\ref{sec:app_proofs}.

\begin{figure}[ht!]
    \centering

    \includegraphics[width=0.22\textwidth]{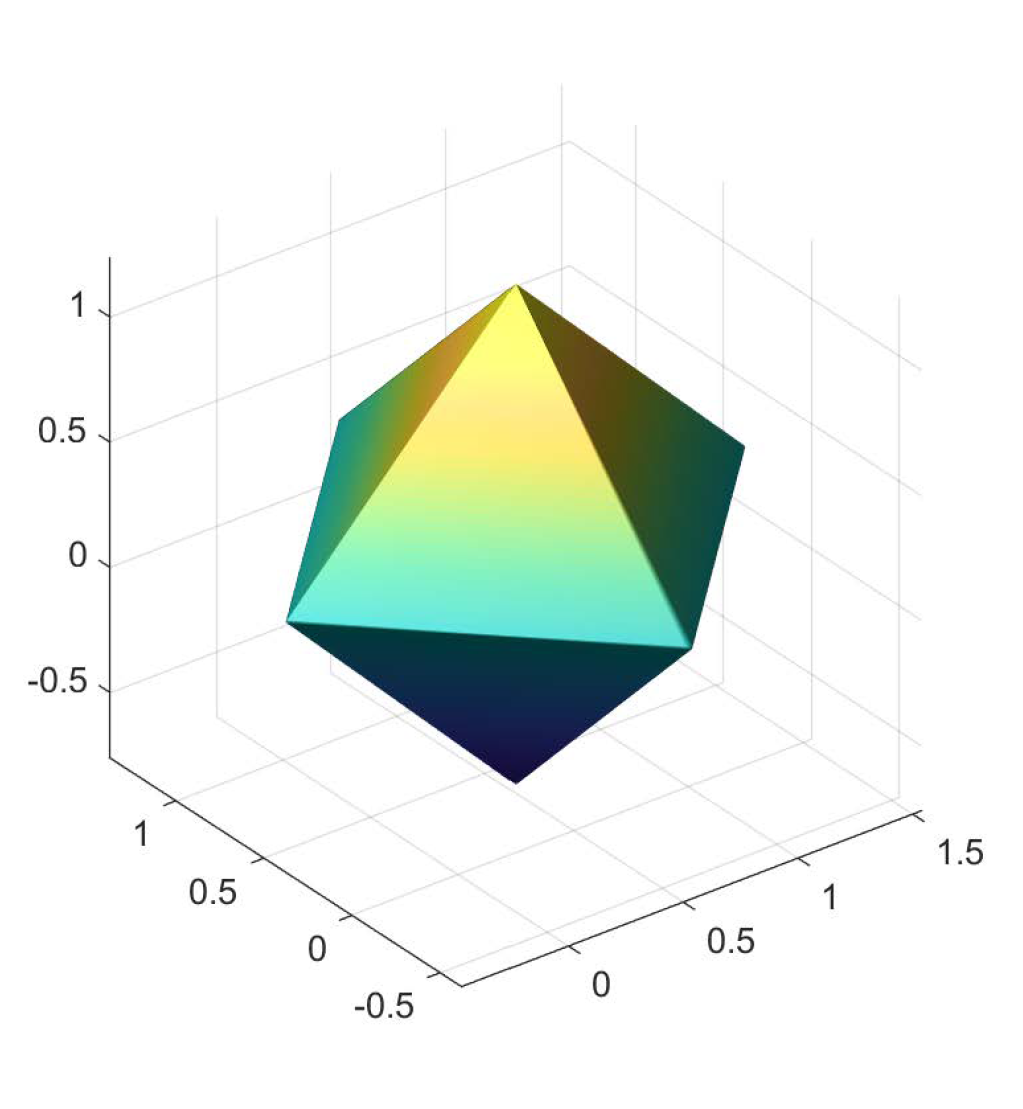}
    \includegraphics[width=0.22\textwidth]{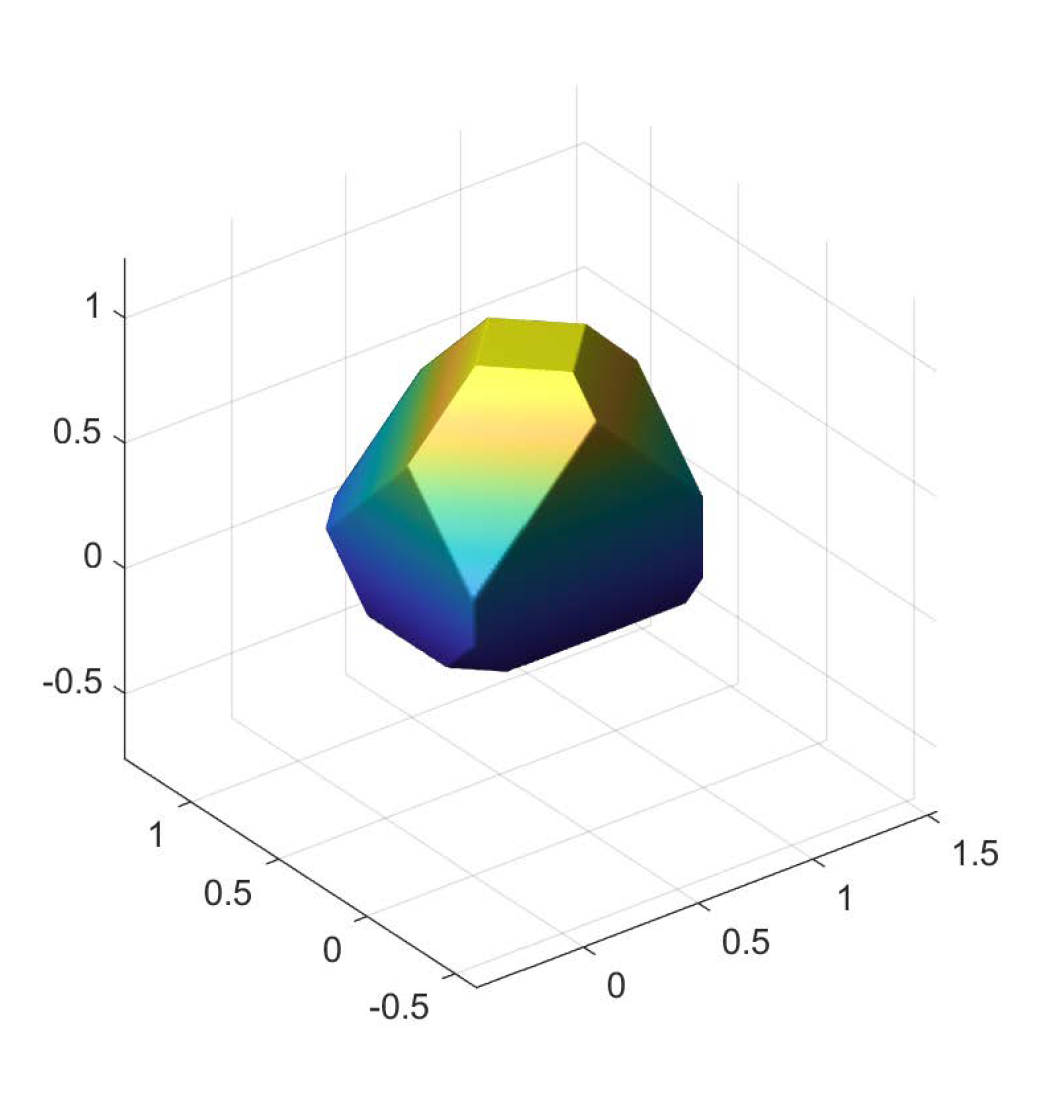}
    \caption{\label{fig:diff}Left: the $l_1$-ball $B_1(x,1)$ centered at the (randomly chosen) target point  $x \in [0,1]^3$, right: the intersection $S=B_1(x,1) \cap [0,1]^3$ of $B_1(x,1)$ 
    with the box $[0,1]^3$.}
\end{figure}
\section{Mind the box $[0,1]^d$ in $l_1$-PGD} \label{sec:pgd_l1}
Projected Gradient Descent is a simple first-order method which consists in a descent step followed by a projection onto the feasible set $S$, that is, given the current iterate $\iter{x}{i}$, the next iterate $\iter{x}{i+1}$ is computed as
\begin{align} \iter{u}{i+1} = &\iter{x}{i} + \iter{\eta}{i} \cdot s(\nabla L(\iter{x}{i})), \label{eq:update_step} \\ \iter{x}{i+1} = &P_S(\iter{u}{i+1}), \label{eq:projection}
\end{align}
where $d$ is the input dimension, $\iter{\eta}{i}>0$ the step size at iteration $i$, $s:\R^d \rightarrow \R^d$ determines the descent direction as a function of the gradient of the loss $L$ at $\iter{x}{i}$ and $P_S:\R^d \rightarrow S$ is the projection on $S$. With an $l_1$-perturbation model of radius $\epsilon$, we denote by $B_1(x,\epsilon):=\{z \in \R^d\,|\, \norm{z-x}_1\leq \epsilon\}$ the $l_1$-ball around
a target point $x \in [0,1]^d$ and define $S=[0,1]^d \cap B_1(x,\epsilon)$. The main difference to prior work is that we take explicitly into account the image constraint $[0,1]^d$. Note that the geometry of the effective threat model is actually quite different from $B_1(x,\epsilon)$ alone, see Figure \ref{fig:diff} for an illustration. In the following we analyse the projection and the descent step in this effective threat model $S$. As $\epsilon$ is significantly higher for $l_1$ (we use $\epsilon=12$ similar to \cite{maini2020adversarial}) as for $l_2$ (standard $0.5$)
and $l_\infty$ (standard $\frac{8}{255}$)
the difference of the intersection with $[0,1]^d$ to the $l_p$-ball alone is most prominent for the $l_1$-case.

\subsection{Projection onto $S$}
With $B_1(x,\epsilon)$ as defined above and denoting $H=[0,1]^d$ the image box, we consider the two projection problems: 
\begin{align}\label{eq:exact-projection}
 P_{S}(u)&=\hl{\argmin}_{z \in \R^d} \norm{u-z}^2_2 \\ \
         &\textrm{ s.th. } \; \norm{z-x}_1 \leq \epsilon, \quad z \in [0,1]^d. \nonumber
\end{align}
and
\begin{equation}\label{eq:l1-projection}
 P_{B_1(x,\epsilon)}(u)=\hl{\argmin}_{z \in \R^d} \norm{u-z}^2_2  \; \textrm{ s.th. } \; \norm{z-x}_1 \leq \epsilon.
\end{equation}
It is well known that the $l_1$-projection problem in \eqref{eq:l1-projection} can be
solved in $O(d\log d)$ \cite{DucEtAl08,condat2016}. We show now that also the exact projection onto $S$ can be computed with the same complexity (after this paper has been accepted we got aware of \cite{wang2019towards} who derived also the form of the solution of \eqref{eq:exact-projection} but provided no complexity analysis or an algorithm to compute it).
\begin{proposition}\label{pro:exact-projection}
The projection problem \eqref{eq:exact-projection} onto $S=B_1(x,\epsilon)\cap H$ can be solved in $O(d \log d)$ with
solution
\[ z^*_i 
= \begin{cases} 1                            & \textrm{ for } u_i \geq x_i \textrm{ and } 0 \leq \lambda_e^* \leq u_i-1\\
                        u_i - \lambda^*_e           & \textrm{ for } u_i \geq x_i \textrm{ and } u_i-1 < \lambda^*_e \leq u_i-x_i\\
                        x_i                         & \textrm{ for } \lambda^*_e > |u_i-x_i|\\
												u_i + \lambda^*_e           &  \textrm{ for } u_i \leq x_i \textrm{ and } -u_i < \lambda^*_e \leq x_i-u_i\\
												0                           &  \textrm{ for } u_i \leq x_i \textrm{ and }  0 \leq \lambda^*_e \leq -u_i\\
												\end{cases},\]
												where $\lambda^*_e \geq 0$. With $\gamma \in \R^d$ defined as
\[ \gamma_i=\max\{-x_i \mathrm{sign}(u_i-x_i),(1-x_i)\mathrm{sign}(u_i-x_i)\},\]							
it holds 	$\lambda_e^*=0$ if $\sum_{i=1}^d \max\{0,\min\{|u_i-x_i|,\gamma_i\}\leq \epsilon$  and otherwise $\lambda^*_e$
is the solution of 
\[ \sum_{i=1}^d \max\big\{0,\min\{|u_i-x_i|-\lambda^*_e,\gamma_i\}\big\}=\epsilon.\]
\end{proposition}
The two prior versions of PGD  \cite{TraBon2019,maini2020adversarial} for the $l_1$-threat model use the approximation $A:\R^d \rightarrow S$
\[ A(u) = (P_H \circ P_{B_1(x,\epsilon)})(u),\]
instead of the exact projection $P_S(u)$ (see the appendix for a proof that $A(u) \in S$ for any $u \in \R^d$). However, it turns out that the approximation $A(u)$ ``hides'' parts of $S$ due to the following property. 
\begin{lemma}\label{le:proj-properties}
It holds for any $u \in \R^d$,
\[  \norm{P_{S}(u) - x}_1 \geq \norm{A(u)-x}_1.\]
In particular, if $P_{B_1(x,\epsilon)}(u) \notin H$
and $\norm{u-x}_1>\epsilon$ 
and one of the following conditions holds
\begin{itemize}
    \item $\norm{P_S(u)-x}_1=\epsilon$
    \item $\norm{P_S(u)-x}_1<\epsilon$ and $\exists u_i \in [0,1]$ with $u_i\neq x_i$
\end{itemize}
then
\[  \norm{P_S(u) - x}_1 > \norm{A(u)-x}_1.\]
\end{lemma}

\begin{figure*}
    \centering
    \begin{tabular}{c|c}
    \includegraphics[width=0.48\textwidth]{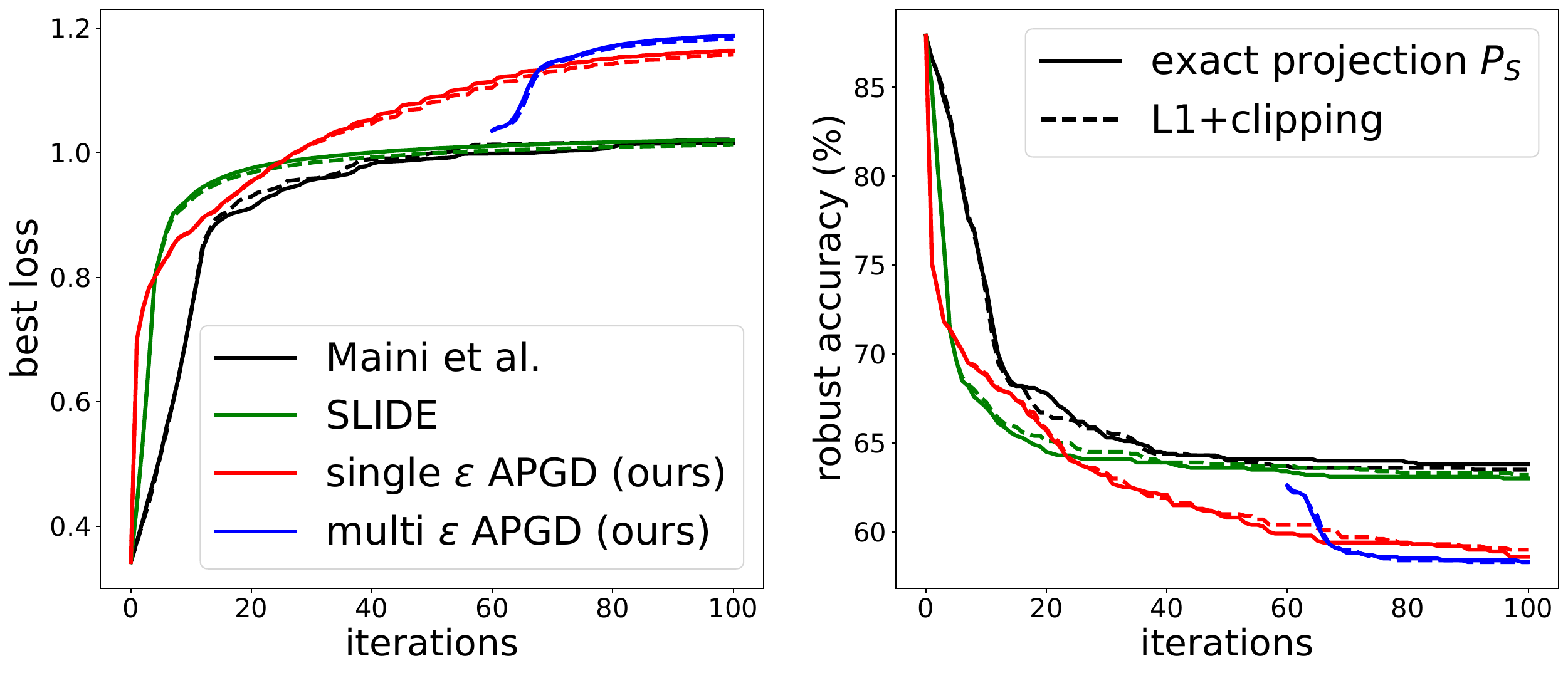} &
    \includegraphics[width=0.48\textwidth]{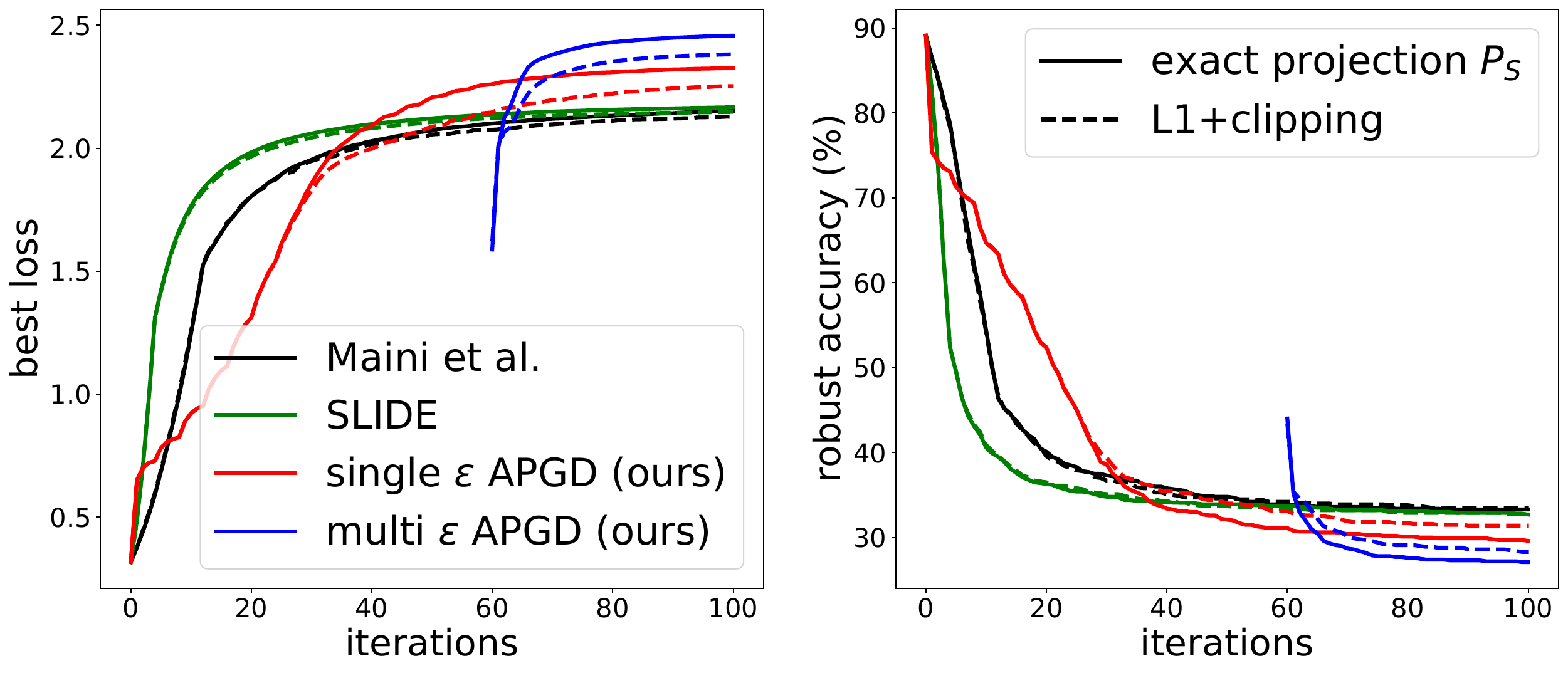}
    \end{tabular}
    \caption{\label{fig:loss-roberror-PGD}Plots of the best robust loss obtained so far (first/third) and robust accuracy (second/fourth) as a function of the iterations for the $l_1$-PGD of \cite{TraBon2019} (SLIDE), the one of \cite{maini2020adversarial} and our single $\epsilon$-APGD and multi $\epsilon$-APGD for two models (left: our own $l_1$-robust model APGD-AT, right: the $l_2$-robust model of \cite{rice2020overfitting}). All of them are once run with the correct projection $P_S(u)$ (solid) and once with the approximation $A(u)$ (dashed). The exact projection improves in almost all cases for all attacks loss and robust accuracy. Moreover, our single- and multi-$\epsilon$ APGD improve significantly the robust loss as well as robust accuracy over SLIDE and the $l_1$-PGD of \cite{maini2020adversarial}. Multi $\epsilon$-APGD is only partially plotted as only the  last $40\%$ of iterations are feasible.}
\end{figure*}
The previous lemma shows that the approximation $A(u)$ of $P_S(u)$ used
by \cite{maini2020adversarial,TraBon2019} is definitely suboptimal under relatively weak conditions and has a smaller $l_1$-distance to the target point $x$:
\[  \norm{P_S(u) - x}_1 > \norm{A(u)-x}_1.\]
Effectively, a part of $S$ is hidden from the attack when $A(u)$ instead of $P_{S}(u)$ is used (see Figure~\ref{fig:ProjRatio} in App.~\ref{sec:app_proofs} for a practical example of this phenomenon). This in turn leads to suboptimal performance both in the maximization of the loss which is important for adversarial training but also in terms of getting low robust accuracy: the plots in Figure \ref{fig:loss-roberror-PGD} show the performance of PGD-based attacks with $A(u)$ (dashed line) vs the same methods with the correct projection $P_S(u)$ (solid line). Our proposed $l_1$-APGD largely benefits from using $P_S(u)$ instead of $A(u)$, and this even slightly improves the existing $l_1$-versions of PGD, SLIDE
\cite{TraBon2019} and the one of \cite{maini2020adversarial}. 
Thus we use in our scheme always the correct projection onto $S$ (in the appendix more statistics on the difference of $A(u)$ and $P_S(u)$).

\subsection{Descent direction}
The next crucial step in the PGD scheme in \eqref{eq:update_step} is the choice of the descent direction which we wrote as the mapping $s(\nabla f(x_i))$ of the gradient.
For the $l_\infty$- and $l_2$-threat models \cite{MadEtAl2018} the steepest descent direction \cite{BoyVan} is used in PGD, that is 
\begin{align} \delta^*_p=\argmax_{\delta\in\R^d} \inner{w, \delta} \quad \text{s.th.} \quad \norm{\delta}_p \leq \epsilon,\label{eq:lin_opt} \end{align} with $w=\nabla f(\iter{x}{i})\in \R^d$, which maximizes a linear function over the given $l_p$-ball. Thus one gets $\delta^*_\infty = \epsilon \sign(w)$ and $\delta^*_2=\epsilon w / \norm{w}_2$ for $p=\infty$ and $p=2$ respectively, which define the function $s$ in \eqref{eq:update_step}. For $p=1$, defining $j=\argmax_{i}{|w_i|}$ the dimension corresponding to the component of $w$ with largest absolute value and $\B=\{e_i\}_i$ the standard basis of $\R^d$, we have $\delta^*_1 = \epsilon \sign(w_j) e_j$. Obviously, for a small number of iterations this descent direction is not working well and thus in SLIDE \cite{TraBon2019} suggest to use the top-$k$ components of the gradient (ordered according to their magnitude) and use the sign of these components.

In the following we show that when one takes into account the box-constraints imposed by the image domain the steepest descent direction becomes automatically less sparse and justifies at least partially what has been done in SLIDE \cite{TraBon2019} and \cite{maini2020adversarial} out of efficiency reasons. 
More precisely, the following optimization problem defines the steepest descent direction:
\begin{align}\label{eq:lin_opt_withbox} \delta^*&= \argmax_{\delta\in\R^d} \inner{w, \delta}\\ & \text{s.th.} 
\quad \norm{\delta}_1 \leq \epsilon, \quad x+\delta \in [0, 1]^d.\nonumber 
\end{align}
\begin{proposition} Let $z_i = \max\{(1-x_i)\sign(w_i), -x_i\sign(w_i) \}$, $\pi$ the ordering such that $|w_{\pi_i}| \geq |w_{\pi_j}|$ for $i > j$ and $k$ the smallest integer for which $\sum_{i=1}^k z_{\pi_i} \geq \epsilon$, then the solution of \eqref{eq:lin_opt_withbox} is given by \begin{align}\delta^*_{\pi_i} = 
\begin{cases} z_{\pi_i} \cdot \sign(w_{\pi_i})& \text{for}\; i < k,\\ (\epsilon - \sum_{i=1}^{k-1} z_{\pi_i}) \cdot \sign(w_{\pi_k})&\text{for}\;i=k, \\ 0 & \text{for}\; i > k\end{cases}. \end{align}\label{prop:sol_lin_with_box} \end{proposition}
Proposition~\ref{prop:sol_lin_with_box} shows that adding the box-constraints leads to a steepest descent direction $\delta^*$ of sparsity level $k$ which depends on the gradient direction $w$ and the target point $x$. Figure~\ref{fig:SparsityLevel} in App.~\ref{sec:app_proofs} provides an empirical evaluation of the distribution of the sparsity level $\norm{\delta^*}_0$.
The following proposition computes the expected sparsity 
of the steepest descent step $\delta^*$
for $\epsilon\leq \frac{d-1}{2}$, together with a simple lower bound.
\begin{proposition}\label{prop:expected_sparsity}
Let $w \in \R^d$ with $w_i\neq 0$ for all $i=1,\ldots,d$ and $x \in \U([0,1]^d)$. Then 
it holds for any $\frac{d-1}{2}\geq \epsilon>0$, 
\begin{align*}
\Exp\big[\norm{\delta^*}_0\big] \hspace{-0.6mm}
=&\hspace{-0.6mm}  \lfloor \epsilon+1 \rfloor + \hspace{-1.5mm}\sum_{m=
\lfloor \epsilon \rfloor + 2}^d \sum_{k=0}^{\lfloor \epsilon \rfloor} (-1)^k \frac{(\epsilon-k)^{m-1}}{k! \, (m-1-k)!} \\ &
\geq \frac{\lfloor 3 \epsilon \rfloor + 1}{2}.
\end{align*}
\end{proposition}
While the exact expression is hard to access, the derived lower bound $\frac{\lfloor 3 \epsilon \rfloor -1}{2}$ shows that the sparsity is non-trivially bounded away from $1$. For a reasonable range of $\epsilon$ the expectation is numerically larger than $2\epsilon$. In this way we provide a justification for the heuristic non-sparse update steps used in \cite{TraBon2019,maini2020adversarial}.

Finally, in our PGD scheme given $g=\nabla L(\iter{x}{i})$, $t\in\N$ and $T(t)$ the set of indices of the $t$ largest components of $|g|$, we define the function $s$ used in \eqref{eq:update_step} via \begin{align} h(t)_i = \begin{cases} \sign(g_i) & \text{if}\, i \in T(t)\\ 0 & \text{else}\end{cases}, \; s(g, t) = h(t)/ \norm{h(t)}_1 \label{eq:sparse_update_step}
\end{align} defines the function $s$ used in \eqref{eq:update_step}. 
The form of the update is the same as in SLIDE \cite{TraBon2019} who use a fixed $k$. However, as derived above, the sparsity level $k$ of the steepest descent direction depends on $\nabla f$ and $x$ and thus we choose $k$ our scheme in a dynamic fashion depending on the current iterate, as described in the next section.

\section{$l_1$-APGD minds $[0,1]^d$}

\begin{algorithm}[tb]
  \caption{Single-$\epsilon$ APGD}
  \label{alg:apgd}
\begin{algorithmic}[1]
  \STATE {\bfseries Input:} loss $L$, initial point $x_\text{init}$, feasible set $S$, \niter{}, $\iter{\eta}{0}$, $\iter{k}{0}$, checkpoints $M$, input dimension $d$
  \STATE {\bfseries Output:} approximate maximizer of the loss $x_\text{best}$
  
  \STATE $\iter{x}{0} \gets x_\text{init}$,
  
  $x_\text{best} \gets x_\text{init}$, $L_\text{best} \gets L(x_\text{init})$
  \FOR{$i=0$ {\bfseries to} \niter{} - 1}
    \STATE \hfill \texttt{// adjust sparsity and step size}
  \IF{$i + 1\in M$}
  \STATE $\iter{k}{i + 1} \gets $ sparsity as in Eq.~\eqref{eq:sparsity_selection}
  \STATE $\iter{\eta}{i + 1} \gets $ step size as in Eq.~\eqref{eq:step_size_selection}
  \IF{$\iter{\eta}{i + 1} = \iter{\eta}{0}$}
        \STATE $\iter{x}{i} \gets x_\text{best}$
  \ENDIF
  \ENDIF

  \STATE \hfill \texttt{// update step} 
  \STATE $\iter{u}{i+1} = \iter{x}{i} + \iter{\eta}{i} \cdot s(\nabla L(\iter{x}{i}), \iter{k}{i + 1} \cdot d)$
  \STATE $\iter{x}{i+1} = P_S(\iter{u}{i+1})$
  \STATE \hfill \texttt{// update best point found}
  \IF{$L(\iter{x}{i+1}) > L_\text{best}$}
  \STATE $x_\text{best} \gets \iter{x}{i+1}$, $L_\text{best} \gets L(\iter{x}{i+1})$
  
  \ENDIF
  \ENDFOR
  
\end{algorithmic} \end{algorithm}

The goal of our $l_1$-APGD is similar to that of APGD for $l_2/l_\infty$ in \cite{croce2020reliable}. It should be parameter-free for the user and adapt the trade-off between exploration and local fine-tuning to the given budget of iterations. 

Proposition~\ref{prop:sol_lin_with_box} suggests the form of the steepest descent direction for the $l_1\text{-ball} \cap [0, 1]^d$ threat model, which has an expected sparsity on the order of $2 \epsilon$ but the optimal sparsity depends on the target point and the gradient of the loss. Thus fixing the sparsity of the update independent of the target point as in SLIDE \cite{TraBon2019} is suboptimal. In practice we have $\epsilon \ll d$ (e.g. for CIFAR-10 $d=3072$ and commonly $\epsilon=12$) and thus $2\epsilon$ sparse updates would lead to slow progress which is in strong contrast to the tight iteration budget used in adversarial attacks. 
Thus we need a scheme where the sparsity is adaptive to the chosen budget of iterations and depends on the current iterate. This motivates two key choices in our scheme: 1) we start with updates with low sparsity, 
2) the sparsity of the updates is then progressively reduced and adapted to the sparsity of the difference of our  currently best iterate (highest loss) to the target point. Thus, initially many coordinates are updated fostering fast progress and exploration of the feasible set, while later on we have a more local exploitation with significantly sparser updates. We observe that, although we do not enforce this actively, the average (over points) sparsity of the updates selected by our adaptive scheme towards the final iterations is indeed on the order of $2\epsilon$, i.e. around to what theoretically expected, although the exact value varies across models. In the following we describe the details of our $l_1$-APGD, see Algorithm \ref{alg:apgd}, and a multi-$\epsilon$ variant which increases the effectiveness, and finally discuss its use for adversarial training  \cite{MadEtAl2018}.

\begin{figure*}[t]
    \centering
    \begin{tabular}{c|c}
    \includegraphics[width=0.48\textwidth]{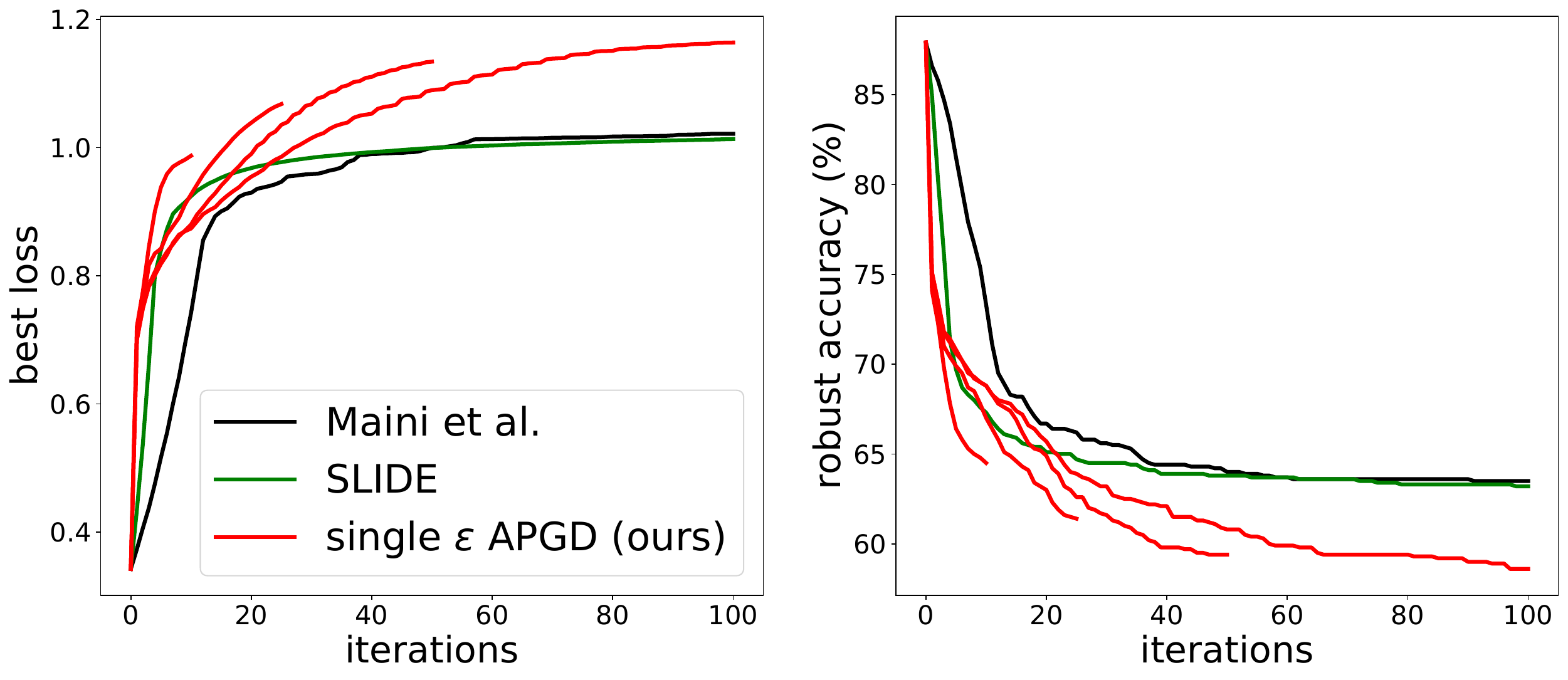} &\includegraphics[width=0.48\textwidth]{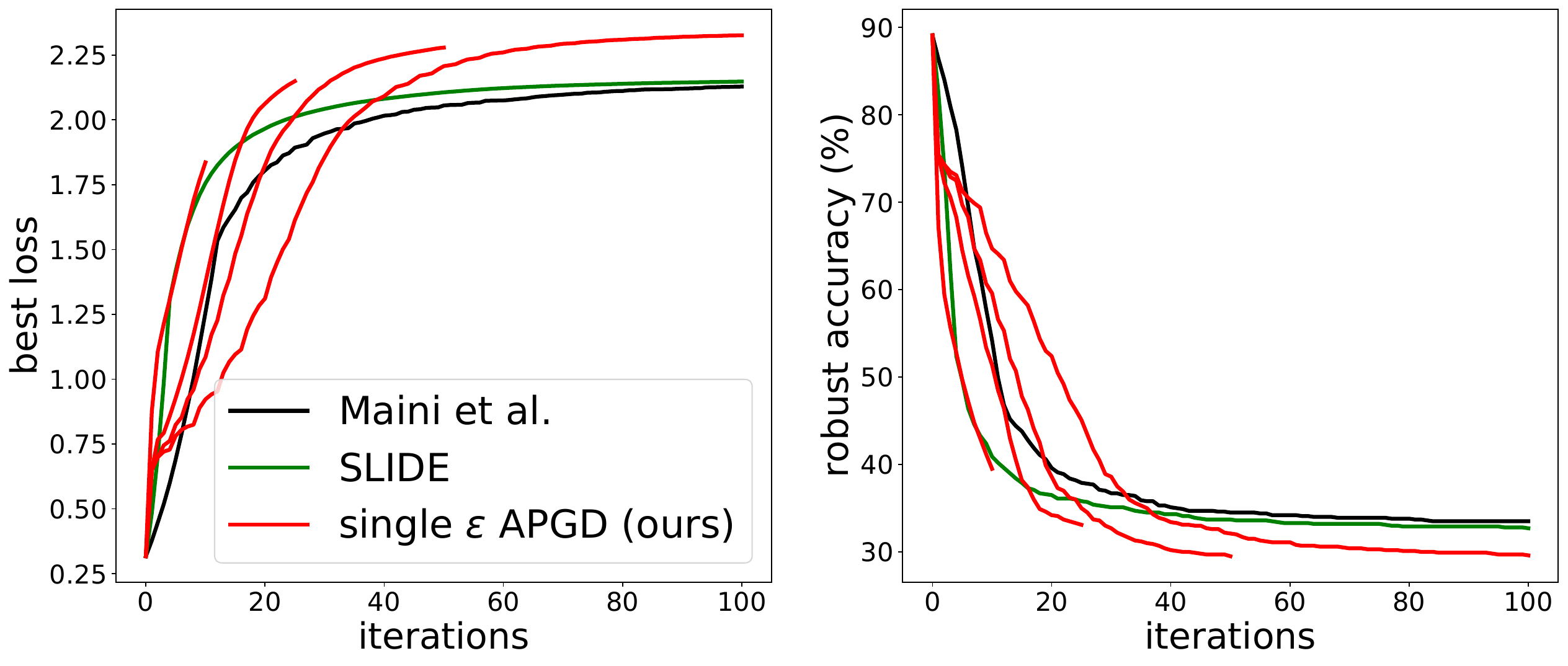}
\end{tabular}
\caption{\label{fig:different_iterations}Plots of best robust loss obtained so far (first/third) and robust accuracy (second/fourth) over iterations for the $l_1$-PGD of \cite{TraBon2019} (SLIDE), the one of \cite{maini2020adversarial} and our single-$\epsilon$ $l_1$-APGD with 10 (the version used for adversarial training), 25, 50 and 100 iterations 
for two models (left: our own $l_1$-robust model, right: the $l_2$-robust one of \cite{rice2020overfitting}). Since our method relies on an adaptive scheme, it automatically adjusts the parameters to the number of available iterations, outperforming the competitors.}
\end{figure*}

\subsection{Single-$\epsilon$ $l_1$-APGD} \label{sec:single_eps_apgd}
Our scheme should automatically adapt to the total budget of iterations. Since the two main quantities which control the optimization in the intersection of $l_1$-ball and $[0,1]^d$ are the sparsity of the updates and the step size, we propose 
to adaptively select them at each iteration.
In particular, we adjust both every $m = \lceil 0.04 \cdot N_\text{iter}\rceil$ steps, with \niter{} being the total budget of iterations, so that every set of parameters is applied for a minimum number of steps to achieve improvement. In the following we denote by $\iter{x_\text{max}}{i}$ the point attaining the highest loss found until iteration $i$, and by $M = \{ n \in \N\, |\,n \mod m = 0\}$ the set of iterations at which the parameters are recomputed.

\textbf{Selection of sparsity:}
We choose an update step for $l_1$-APGD whose sparsity is automatically computed by considering the best point found so far.
In order to have both sufficiently fast improvements and a good exploration of the feasible set in the first iterations of the algorithm, we start with updates $d / 5$ with nonzero elements, that is a sparsity $\iter{k}{0}=0.2$ (in practice this implies $\iter{k}{0} \gg 2\epsilon / d$). Then, the sparsity of the updates is adjusted as
\begin{align}
\iter{k}{i} = \begin{cases} \norm{\iter{x_\text{max}}{i-1} - x}_0 / (1.5\cdot d) & \text{if}\; i \in M, \\ \iter{k}{i - 1} & \text{else}.
\end{cases}\label{eq:sparsity_selection} \end{align}

Note that $\iter{k}{i}$ is smaller than the sparsity of the current best perturbation since the initial perturbations have much larger $l_0$-norm than the expected $2 \epsilon$, and we want to refine them.

\textbf{Selection of the step size:} Simultaneously to the sparsity of the updates, we adapt the step size to the trend of the optimization. As high level idea:  if the $l_0$-norm of the best solution is not decreasing significantly for many iterations, this suggests that it is close to the optimal value and that the step size is too large to make progress, then we reduce it. Conversely, when the sparsity of the updates keeps increasing we want to allow large step sizes since the region of the feasible set which can be explored by sparser updates is different from what can be seen with less sparse steps. We set the initial step size $\iter{\eta}{0} = \epsilon$ (the radius of the $l_1$-ball), so that the algorithm can search efficiently the feasible set, and then adjust the step size at iterations $i\in M$ according to 
\begin{align} \iter{\eta}{i} = \begin{cases} \max\{\iter{\eta}{i - m} / 1.5, \eta_\text{min}\} & \text{if}\; 
\iter{k}{i}/\iter{k}{i-m} \geq 0.95,
\\
\iter{\eta}{0} & 
\text{else},
\end{cases} \label{eq:step_size_selection} \end{align}
where $\eta_\text{min} = \epsilon / 10$ is the smallest value we allow for the step size.
Finally, when the step size is set to its highest value, the algorithm restarts from $\iter{x_\text{max}}{i}$.

\subsection{Multi-$\epsilon$ $l_1$-APGD} \label{sec:multi_eps_apgd}
In the described $l_1$-APGD all iterates belong to the feasible set $S$. However, since the points maximizing the loss are most likely on the low dimensional faces of $S$, finding them might require many iterations. We notice that the same points are instead in the interior of any $l_1$-ball with radius larger than $\epsilon$. Thus we propose to split \niter{} into three phases with 30\%, 30\% and 40\% of the iteration budget, where we optimize the objective $L$ in $l_1$-balls of radii $3\epsilon$, $2 \epsilon$ and $\epsilon$ (always intersected with $[0,1]^d$) respectively. At the beginning of each phase the output of the previous one is projected onto the intersection of the next $l_1$-ball and $[0,1]^d$ and used as starting point for $l_1$-APGD with smaller radius. In this way we efficiently find good regions where to start the optimization in the target feasible set, see Figure \ref{fig:loss-roberror-PGD} for an illustration.

\subsection{Comparison PGD vs single-$\epsilon$ and multi-$\epsilon$ $l_1$-APGD}
In Figure~\ref{fig:loss-roberror-PGD} we compare the performance of the versions of PGD used by SLIDE \cite{TraBon2019} and \cite{maini2020adversarial} to our $l_1$-APGD, in both single- and multi-$\epsilon$ variants. For two models on CIFAR-10, we plot the best average (over 1000 test points) cross-entropy loss achieved so far and the relative robust accuracy (classification accuracy on the adversarial points), with a total budget of 100 iterations. For multi-$\epsilon$ APGD, we report the results only from iteration 60 onward, since before that point the iterates are outside the feasible set and thus the statistics not comparable. We see that the single-$\epsilon$ APGD achieves higher (better) loss and lower (better) robust accuracy than the existing PGD-based attacks, which tend to quickly plateau. Also, multi-$\epsilon$ APGD provides an additional improvement: exploring the larger $l_1$-ball yields an initialization in $S$ with high loss, from where even a few optimization steps are sufficient to outperform the other methods. Additionally, we observe that using the exact projection (solid lines) boosts the effectiveness of the attacks compared to the approximated one (dashed lines), especially on the $l_2$-robust model of \cite{rice2020overfitting} (the two rightmost plots).
Moreover, we show in Figure~\ref{fig:different_iterations} how our single-$\epsilon$ APGD adapts to different budgets of iterations: when more steps are available, the loss improves more slowly at the beginning, favoring the exploration of the feasible set, but it finally achieves better values. Note that in this way, even with only 25 steps, $l_1$-APGD outperforms existing methods with 100 iterations both in terms of loss and robust accuracy attained.

\subsection{Adversarial training with $l_1$-APGD} \label{sec:adv_train}
A natural application of a strong PGD-based attack is to maximize the loss in the inner maximization problem 
of adversarial training (AT) \cite{MadEtAl2018} and finding points attaining higher loss should lead to more adversarially robust classifiers. 
Prior works have shown that performing adversarial training wrt $l_1$ is a more delicate task than for other $l_p$-threat models: for CIFAR-10, \cite{maini2020adversarial} report that using PGD wrt $l_1$ in AT led to severe gradient obfuscation, and the resulting model is less than 8\% robust at $\epsilon=12$. A similar effect is reported in \cite{liu2020towards}, where the B\&B attack of \cite{brendel2019accurate} more than halves 
the robust accuracy computed by $l_1$-PGD used for their $l_1$-AT model. Both report the highest robustness to $l_1$-attacks when training for simultaneous robustness against different $l_p$-norms.
In our own experiment, which are discussed in more details in App.~\ref{sec:app_adv_training}, we observed that AT wrt $l_1$, even with multi-step PGD, is prone to catastrophic overfitting (CO) as described by \cite{Wong2020Fast}. Thus, even if the AT training seemingly works, the resulting classifier is still non-robust, suggesting some kind of overfitting to the adversarial samples generated by $l_1$-PGD. While we have no final explanation for this, our current hypothesis is that standard $l_1$-PGD produces adversarial samples of a certain sparsity level with little variation and thus the full threat model is not explored during training. 
In contrast, $l_1$-APGD, which progressively and adaptively adjusts the sparsity, mitigates the risk of CO while providing strong adversarial perturbations. We leave it to future work to do a more thorough investigation of this interesting phenomenon.
We apply $l_1$-APGD (single-$\epsilon$ formulation with initial sparsity $\iter{k}{0} = 0.05$) with 10 steps to train a ResNet-18 (details in App.~\ref{sec:app_exps}) for an $l_1$-threat model with radius $\epsilon=12$.
In Sec.~\ref{sec:exp} we show that our APGD-AT model achieves significantly higher robust accuracy than the currently best model, even in the worst case evaluation over many strong attacks, including black-box ones.

\section{$l_1$-AutoAttack} \label{sec:l1_aa}
\cite{croce2020reliable} propose AutoAttack (AA), an ensemble of four diverse attacks for a standardized parameter-free and reliable evaluation of robustness against $l_\infty$- and $l_2$-type attacks, and we aim to extend this framework to the case of $l_1$-robustness.
AA includes the $l_\infty$- and $l_2$-APGD optimizing either the cross-entropy (CE) or targeted version of the difference of logits ratio (T-DLR) loss \cite{croce2020reliable}: analogously we use our multi-$\epsilon$ $l_1$-APGD with 5 runs (with random restarts) of 100 iterations for the CE and the T-DLR loss (total budget of 1000 steps).
The targeted FAB-attack included in AA \cite{CroHei2019}  minimizes the norm of the adversarial perturbations and has an $l_1$-version, therefore no action is needed (run with top 9 classes). The black-box Square Attack \cite{ACFH2019square} has only versions for $l_\infty$- and $l_2$-bounded perturbations, hence we adapt the latter to the $l_1\cap [0,1]^d$-threat model (details in Sec.~\ref{sec:square}). We show in the experiments that having a black-box method helps to accurately estimate robustness even in presence of defenses with gradient obfuscation. For both FAB\SP{T} and Square Attack (5000 queries) we keep the budget of iterations and restarts defined in AA for $l_\infty$ and $l_2$. Note that the parameters of all attacks are fixed so that no tuning is necessary when testing different models and thus we get a parameter-free 
$l_1$-AutoAttack which achieves SOTA performance as we show in Section \ref{sec:exp}.

\subsection{$l_1$-Square Attack} \label{sec:square}
\cite{ACFH2019square} introduce Square Attack, a query efficient score-based black-box adversarial attack for $l_\infty$- and $l_2$-bounded perturbations. It is based on random search and does not rely on any gradient estimation technique. \cite{ACFH2019square} show that it does not suffer from gradient masking and is even competitive with white-box attacks in some scenarios. We adapt its $l_2$ version to our $l_1$-threat model, by modifying Algorithm~3 in the original paper so that all normalization operations are computed wrt the $l_1$-norm (see App.~\ref{sec:app_square} for details). 
While \cite{ACFH2019square} create at every iteration perturbations on the surface of the $l_p$-ball and then clip them to $[0,1]^d$, this results in poor performance for the $l_1$-ball, likely due to the complex structure of the intersection of $l_1$-ball and $[0,1]^d$ (see discussion above). Thus, at each iteration, we upscale the square-shaped candidate update by a factor of $3$ 
and then project the resulting iterate onto the intersection of the $l_1$-ball and $[0,1]^d$ and accept this update if it increases the loss. 
This procedure increases the sparsity of the iterates and in turn the effectiveness of the attack, showing again the different role that the box has in the $l_1$-threat model compared to the $l_\infty$- and $l_2$-threat models. We show 
below that our resulting scheme, $l_1$-Square Attack, outperforms the existing black-box methods \citep{SchEtAl19,zhao2019design} on a variety of models, often with margin.

\begin{table*}[t] \caption{\label{tab:cheap}\textbf{Low Budget (\boldmath $\epsilon=12$\unboldmath):} Robust accuracy achieved by the SOTA $l_1$ -adversarial attacks on various models for CIFAR-10 in the $l_1$-threat model with radius $\epsilon=12$ of the $l_1$-ball. The statistics are computed on 1000 points of the test set. PA and Square are black-box attacks. The budget is 100 iterations for white-box attacks ($\times$9 for EAD and +10 for B\&B) and $5000$ queries for our $l_1$-Square-Attack.} \centering \vspace{2mm} {\small
\setlength{\tabcolsep}{1.5pt}
\begin{tabular}{L{40mm} | C{12mm} | *{5}{C{12mm}} >{\columncolor[rgb]{0.9 0.9 0.9}}C{12mm} |C{12mm}>{\columncolor[rgb]{0.9 0.9 0.9}}C{12mm}
}
\textit{model} & clean &EAD & ALMA & SLIDE & B\&B\SPSB{}{} & FAB\SPSB{T}{} & APGD\SPSB{}{CE} & PA & Square\\ \toprule  APGD-AT \textbf{(ours)} & 87.1& 64.6& 65.0& 66.6& 62.4& 67.5& \textbf{61.3}& 79.7& 71.8\\ \cite{madaan2020learning} & 82.0& 55.3& 58.1& 56.1& 55.2& 56.8& \textbf{54.7}& 73.1& 62.8\\ \cite{maini2020adversarial} - AVG & 84.6& 51.8& 54.2& 53.8& 52.1& 61.8& \textbf{50.4}& 77.4& 68.4\\ \cite{maini2020adversarial} - MSD & 82.1& 51.6& 55.4& 53.2& 50.7& 54.6& \textbf{49.7}& 72.7& 63.5\\ \cite{augustin2020} & 91.1& 48.9& 50.7& 48.8& 42.1& 50.4& \textbf{37.1}& 73.2& 56.8\\ \cite{robustness} - $l_2$ & 91.5& 40.3& 46.4& 35.1& 36.8& 39.9& \textbf{30.2}& 71.7& 52.7\\ \cite{rice2020overfitting} & 89.1& 37.7& 45.2& 32.3& 35.2& 37.0& \textbf{27.1}& 70.5& 50.3\\ \cite{xiao2020enhancing} & 79.4& 44.9& 74.5& 33.3& 72.6& 78.9& 41.4& 36.2& \textbf{20.2}\\ \cite{kim2020adversarial}\SP{*} & 81.9& 26.7& 31.8& 25.1& 23.8& 32.4& \textbf{18.9}& 54.9& 36.0\\ \cite{CarEtAl19} & 90.3& 25.1& 18.4& 19.7& 18.7& 31.1& \textbf{13.1}& 60.8& 34.5\\ \cite{xu2021adversarial} & 83.8& 20.1& 24.0& 18.2& 14.7& 27.8& \textbf{10.9}& 57.0& 32.0\\ \cite{robustness} - $l_\infty$ & 88.7& 14.5& 19.4& 14.2& 12.2& 20.9& \textbf{8.0}& 57.6& 28.0
\\ \bottomrule
\end{tabular}} \end{table*}

\begin{table*}[h]\caption{\label{tab:expensive}\textbf{High Budget (\boldmath $\epsilon=12$\unboldmath):} Robust accuracy achieved by the SOTA $l_1$ -adversarial attacks on various models for CIFAR-10 in the $l_1$-threat model with $l_1$-radius of $\epsilon=12$. The statistics are computed on 1000 points of the test set. ``WC'' denotes the pointwise worst-case over all restarts/runs of EAD, ALMA, SLIDE, B\&B and Pointwise Attack. Note that APGD\SPSB{}{CE+T}, the combination of APGD\SPSB{}{CE} and  APGD\SPSB{}{T-DLR} ($5$ restarts each), yields a similar performance as AA (ensemble of APGD\SPSB{}{CE+T}, $l_1$-FAB$^T$ and $l_1$-Square Attack)  with the same or smaller budget than the other individual attacks. AA performs the same or beats the worst case WC of five SOTA $l_1$-attacks in 8 out of 12 cases. ``rep.'' denotes the reported robust accuracy in the original papers.
$^{*}$ the models of \cite{kim2020adversarial} were not available on request and thus are retrained with their code (see appendix).
$^{**}$ In  \cite{madaan2020learning} evaluation is done at $\epsilon=\frac{2000}{255}$, but by personal communication with the authors we found that the reported $55.0\%$ corresponds to $\epsilon=12$.} \vspace{2mm}
\centering
{\centering \small 
\tabcolsep=1.8pt
\begin{tabular}{L{40mm} | C{13mm} | *{4}{C{13mm}} >{\columncolor[rgb]{0.9 0.9 0.9}}C{13mm}| C{13mm} >{\columncolor[rgb]{0.9 0.9 0.9}}C{13mm}| C{13mm}}
\textit{model}&
clean &EAD & ALMA & SLIDE & B\&B\SPSB{}{} & APGD\SPSB{}{CE+T} & WC & AA& rep.\\ \toprule  APGD-AT \textbf{(ours)} & 
87.1& 63.3& 61.4& 65.9& 59.9& 60.3& \textbf{59.7}& 60.3&-\\ \cite{madaan2020learning} & 82.0& 54.5& 54.3& 55.1& 51.9& 51.9& \textbf{51.8}& 51.9& 55.0\SP{**}\\ \cite{maini2020adversarial} - AVG & 84.6& 50.0& 49.7& 52.3& 49.0& \textbf{46.8}& 47.3& \textbf{46.8}& 54.0\\ \cite{maini2020adversarial} - MSD & 82.1& 50.1& 49.8& 51.7& 47.7& \textbf{46.5}& 46.8& \textbf{46.5}& 53.0\\ \cite{augustin2020} & 91.1& 46.0& 42.9& 41.5& 32.9& 31.1& 31.9& \textbf{31.0}& -\\ \cite{robustness} - $l_2$ & 91.5& 36.4& 34.7& 30.6& 27.5& 27.0& 27.1& \textbf{26.9}& -\\ \cite{rice2020overfitting} & 89.1& 33.9& 32.4& 28.1& 24.2& 24.2& \textbf{23.7}& 24.0& -\\ \cite{xiao2020enhancing} & 79.4& 34.4& 75.0& 22.5& 59.3& 27.2& 20.2& \textbf{16.9}& -\\ \cite{kim2020adversarial}\SP{*} & 81.9& 24.4& 22.9& 19.9& 15.7& 15.4& \textbf{15.1}& \textbf{15.1}& 81.18\\ \cite{CarEtAl19} & 90.3& 26.2& 13.6& 13.6& 10.4& \textbf{8.3}& 8.5& \textbf{8.3}& -\\ \cite{xu2021adversarial} & 83.8& 18.1& 14.5& 13.9& 7.8& 7.7& \textbf{6.9}& 7.6& 59.63\\ \cite{robustness} - $l_\infty$ & 88.7& 12.5& 10.0& 8.7& 5.9& \textbf{4.9}& 5.1& \textbf{4.9} & - 
\\ \bottomrule
\end{tabular} } \end{table*}

\begin{table*}[h]\caption{\label{tab:expensive2}\textbf{High Budget (\boldmath $\epsilon=8$\unboldmath):} see Table \ref{tab:expensive} for details, the only change is the evaluation of robust accuracy at the smaller value $\epsilon=8$.}
\vspace{+2mm}
\centering {\small 
\tabcolsep=2pt
\begin{tabular}{L{40mm} | C{14mm} | *{4}{C{14mm}} >{\columncolor[rgb]{0.9 0.9 0.9}}C{14mm}| C{14mm} >{\columncolor[rgb]{0.9 0.9 0.9}}C{14mm}}
\textit{model} & clean &EAD & ALMA & SLIDE & B\&B\SPSB{}{} & APGD\SPSB{}{CE+T} & WC & AA\\ \toprule  APGD-AT \textbf{(ours)} & 
87.1& 71.6& 71.8& 72.9& \textbf{70.6}& \textbf{70.6}& \textbf{70.6}& \textbf{70.6}\\ \cite{madaan2020learning} & 82.0& 62.6& 63.3& 62.7& \textbf{60.6}& 60.7& \textbf{60.6}& \textbf{60.6}\\ \cite{maini2020adversarial} - AVG & 84.6& 62.9& 63.1& 63.1& 62.4& \textbf{60.3}& 61.3& \textbf{60.3}\\ \cite{maini2020adversarial} - MSD & 82.1& 60.2& 61.0& 61.6& 58.6& 58.3& \textbf{58.2}& \textbf{58.2}\\ \cite{augustin2020} & 91.1& 60.9& 60.1& 60.8& 52.6& \textbf{50.7}& 52.0& \textbf{50.7}\\ \cite{robustness} - $l_2$ & 91.5& 54.0& 54.9& 52.3& 46.0& 44.4& 45.9& \textbf{44.2}\\ \cite{rice2020overfitting} & 89.1& 52.5& 51.5& 49.9& 44.3& \textbf{42.9}& 44.1& \textbf{42.9}\\ \cite{kim2020adversarial}\SP{*} & 81.9& 38.7& 38.9& 36.6& 31.8& 30.4& 31.4& \textbf{30.1}\\ \cite{xiao2020enhancing} & 79.4& 41.2& 75.3& 30.7& 60.6& 33.8& 27.9& \textbf{22.4}\\ \cite{CarEtAl19} & 90.3& 37.8& 29.7& 28.8& 25.1& \textbf{21.1}& 22.6& \textbf{21.1}\\ \cite{xu2021adversarial} & 83.8& 33.1& 30.2& 27.6& 22.6& 21.4& 21.6& \textbf{21.0}\\ \cite{robustness} - $l_\infty$ & 88.7& 28.8& 24.9& 23.1& 17.5& 16.5& 16.4& \textbf{16.0}
\\\bottomrule
\end{tabular}} \end{table*}

\section{Experiments} \label{sec:exp}

In the following we test the effectiveness of our proposed attacks.\footnote{Code available at \url{https://github.com/fra31/auto-attack}.} First we compare $l_1$-APGD (with CE loss) and our $l_1$-Square Attack to existing white- and black-box attacks in the low budget regime, then we show that $l_1$-AutoAttack accurately evaluates of robustness wrt $l_1 \cap [0,1]^d$ for all the models considered. All attacks are evaluated on models trained on CIFAR-10 \cite{CIFAR10} and we report robust accuracy for $\epsilon=8$ and $\epsilon=12$ on 1000 test points.

\textbf{Models:} The selected models are (almost all) publicly available and are representative of different architectures and training schemes: the models of \cite{CarEtAl19,robustness,xiao2020enhancing,kim2020adversarial,xu2021adversarial} are robust wrt $l_\infty$, where the model of \cite{xiao2020enhancing} is known to be non-robust but shows heavy gradient obfuscation, those of \cite{augustin2020, robustness, rice2020overfitting} wrt $l_2$, while those of \cite{maini2020adversarial, madaan2020learning} are trained for simultaneous robustness wrt $l_\infty$-, $l_2$- and $l_1$-attacks. To our knowledge no prior work has focused on training robust models for solely $l_1$ (see discussion in Sec. \ref{sec:adv_train}). Additionally, we include the classifier we trained with $l_1$-APGD integrated in the adversarial training of \cite{MadEtAl2018} and indicated as APGD-AT (further details in the appendix).

\textbf{Attacks:} We compare our attacks to the existing SOTA attacks for the $l_1$-threat model using their existing code (see appendix for hyperparameters). In detail, we consider SLIDE \cite{TraBon2019} (an attack based on PGD), EAD \cite{CheEtAl2018}, FAB\SP{T} \cite{CroHei2019}, 
B\&B \cite{brendel2019accurate} and the recent ALMA \cite{rony2020augmented}. As reported in \cite{rony2020augmented} B\&B crashes as the initial procedure to sample uniform noise to get a decision different from the true class fails. 
Thus we initialize B\&B with random images from CIFAR-100 
(the results do not improve when starting at CIFAR-10 images).
Besides white-box methods we include the black-box Pointwise Attack (PA) \cite{SchEtAl19}, introduced for the $l_0$-threat model but successfully used as $l_1$-attack by e.g. \cite{maini2020adversarial}.
We always use our $l_1$-APGD in the multi-$\epsilon$ version.

\emph{Small Budget:} We compare the attacks with a limited computational budget, i.e. 100 iterations, with the exception of EAD for which we keep the default 9 binary search steps (that is $9\times 100$ iterations), B\&B which performs an initial 10 step binary search procedure (10 additional forward passes). Moreover, we add the black-box attacks Pointwise Attack and our $l_1$-Square Attack with 5000 queries (no restarts). Table~\ref{tab:cheap} reports the robust accuracy at $\epsilon=12$ achieved by every attack: in all but one case $l_1$-APGD\SPSB{}{CE} maximizing the cross-entropy loss outperforms the competitors, in 6 out of 11 cases with a gap larger than $4\%$ to the second best method. 
Note that $l_1$-APGD\SPSB{}{CE} consistently achieves lower (better) robustness with a quite significant gap to the non adaptive PGD-based attack SLIDE. Note also that $l_1$-APGD\SPSB{}{CE}, SLIDE and ALMA are the fastest attacks for the budget of 100 iterations (see appendix for more details).  The  model from \cite{xiao2020enhancing} exemplifies the importance of testing robustness also with black-box attacks: their defense generates gradient obfuscation so that white-box attacks have difficulties to perform well (in particular ALMA, FAB\SP{T} and B\&B yield a robust accuracy close to the clean one), while Square Attack is not affected and achieves the best results with a large margin. Moreover, it outperforms on all models the other black-box attack PA.
This supports its inclusion in AutoAttack.

\emph{High budget:} As second comparison, we give the attacks a higher budget: we use SLIDE, FAB\SP{T} and B\&B with 10 random restarts of 100 iterations (the reported accuracy is then the pointwise worst case over restarts). B\&B has a default value of $1000$ iterations but $10$ restarts with $100$ iterations each yield much better results. For ALMA and EAD we use 1000 resp. $9\times 1000$ iterations (note that EAD does a binary search) since they do not have the option of restarts. We compare these strong attacks to the combination of $l_1$-APGD\SPSB{}{CE} and $l_1$-APGD\SPSB{}{T-DLR} denoted as APGD\SPSB{}{CE+T} with $100$ iterations and $5$ restarts each. These runs are also part of our ensemble  $l_1$-AutoAttack introduced in Sec.~\ref{sec:l1_aa} which includes additionally, FAB\SP{T} (100 iterations and 9 restarts as used in $l_\infty$- and $l_2$-AA) and Square Attack (5000 queries). Note that APGD\SPSB{}{CE+T} has the same or smaller budget than the other attacks and it performs very similar to the full $l_1$-AutoAttack (AA). In Table~\ref{tab:expensive} and ~\ref{tab:expensive2} we report the results achieved by all methods for $\epsilon=12$ resp. $\epsilon=8$. AA outperforms for $\epsilon=12$ the individual competitors in all cases except one, i.e. B\&B on the APGD-AT model, where however it is only 0.5\% far from the best. Also, note that all the competitors have at least one case where they report a robust accuracy more than 10\% worse than AA. The same also holds for the case $\epsilon=8$. Note that for $\epsilon=12$ APGD\SPSB{}{CE+T} is the best single attack in 10 out of 12 cases (B\&B 3, SLIDE 1) and for $\epsilon=8$ APGD\SPSB{}{CE+T} is the best in 9 out of 12 cases 
(B\&B 2, SLIDE 1).

Since AA is an ensemble of methods, as a stronger baseline we additionally report the worst case robustness across all the methods not included in AA (indicated as ``WC" in the tables), that is EAD, ALMA, SLIDE, B\&B and Pointwise Attack. 
$l_1$-AA achieves better results in most of the cases (7/12 for $\epsilon=12$, 9/12 for $\epsilon=8$) even though it has less than half of the total budget of WC. AA is 
0.6\% worse than WC for the APGD-AT model (at $\epsilon=12$), which is likely due to the fact that $l_1$-APGD is used to generate adversarial examples at training time and thus there seems to be a slight overfitting effect to this attack. However, note that standard AT with $l_1$-PGD has been reported to fail completely. We report the results of the individual methods of $l_1$-AA in App.~\ref{sec:indiv_attacks}.
We also list in Table~\ref{tab:expensive} the $l_1$-robust accuracy reported in the original papers, if available.
The partially large differences indicate that a standardized evaluation with AA would lead to a more reliable
assessment of $l_1$-robustness.

\begin{table*}[t]
\caption{Comparison of black-box attacks in the $l_1$-threat model on CIFAR-10, $\epsilon=12$. $l_1$-Square Attack outperforms both the pointwise attack \cite{SchEtAl19} and the $l_1$-ZO-ADMM-Attack  \cite{zhao2019design} by large margin. \label{tab:black_box_comp}}
\centering \small \vspace{2mm}
\begin{tabular}{L{40mm} | C{12mm} | C{12mm} C{12mm}  >{\columncolor[rgb]{0.9 0.9 0.9}}C{12mm}}
\textit{model} & clean &ADMM & PA & Square\\ \toprule  APGD-AT \textbf{(ours)} & 
87.1& 86.3& 79.7& \textbf{71.8}\\ \cite{maini2020adversarial} - AVG & 84.6& 81.3& 77.4& \textbf{68.4}\\ \cite{maini2020adversarial} - MSD & 82.1& 77.5& 72.7& \textbf{63.5}\\ \cite{madaan2020learning} & 82.0& 78.4& 73.1& \textbf{62.8}\\ \cite{augustin2020} & 91.1& 88.9& 73.2& \textbf{56.8}\\ \cite{robustness} - $l_2$ & 91.5& 89.8& 71.7& \textbf{52.7}\\ \cite{rice2020overfitting} & 89.1& 85.9& 70.5& \textbf{50.3}\\ \cite{kim2020adversarial}\SP{*} & 81.9& 67.8& 54.9& \textbf{36.0}\\ \cite{CarEtAl19} & 90.3& 64.1& 60.8& \textbf{34.5}\\ \cite{xu2021adversarial} & 83.8& 66.0& 57.0& \textbf{32.0}\\ \cite{robustness} - $l_\infty$ & 88.7& 69.3& 57.6& \textbf{28.0}\\ \cite{xiao2020enhancing} & 79.4& 78.5& 36.2& \textbf{20.2}
\\ \bottomrule
\end{tabular}\end{table*}
The most robust model is APGD-AT trained with our single-$\epsilon$ APGD (see Sec.~\ref{sec:adv_train}) for $\epsilon=12$. 
It improves by 
7.9\% 
over the second best model \cite{madaan2020learning} (see Table~\ref{tab:expensive}). This highlights how effective our $l_1$-APGD maximizes the target loss, even with only 10 steps used in AT.

\textbf{Comparison of black-box attacks:}
While many black-box attacks are available for the $l_\infty$- and $l_2$-threat model, and even a few have recently appeared for $l_0$, the $l_1$-threat model has received less attention: in fact, \cite{TraBon2019,maini2020adversarial} used the Pointwise Attack, introduced for $l_0$, to test robustness wrt $l_1$. To our knowledge only \cite{zhao2019design} have proposed $l_1$-ZO-ADMM, a black-box method to minimize the $l_1$-norm of the adversarial perturbations, although only results on MNIST are reported. Since no code is available for ZO-ADMM for $l_1$, we adapted the $l_2$ version following \cite{zhao2019design} (see App.~\ref{sec:app_exps}). 
As for our $l_1$-Square Attack, we give to $l_1$-ZO-ADMM a budget of 5000 queries of the classifier. Table~\ref{tab:black_box_comp} shows the robust accuracy on 1000 test points achieved by the three black-box attacks considered on CIFAR-10 models, with $\epsilon=12$: Square Attack outperforms the other methods on all models, with a significant gap to the second best. Note that $l_1$-ZO-ADMM does not consider norm-bounded attacks, but minimizes the norm of the modifications. While it is most of the time successful in finding adversarial perturbations, it cannot reduce their $l_1$-norm below the threshold $\epsilon$ within the given budget of queries.

\textbf{Additional experiments:} We include in App.~\ref{sec:app_additional_exps} additional experiments. 
In particular, we study the effect of using different values $k$ of sparsity in SLIDE \cite{TraBon2019}: the default $k=0.01$ achieve the best results on average on CIFAR-10, but the optimal one varies across classifiers (see App.~\ref{sec:app_abl_slide}). This means that using a fixed threshold is suboptimal, and further motivates our adaptive scheme implemented in $l_1$-APGD. 
Moreover, in App.~\ref{sec:other_datasets} we extend the evaluation to other datasets, CIFAR-100 and ImageNet-1k, with a similar setup as above: on both datasets, $l_1$-APGD significantly outperforms the competitors in the low budget regime, and $l_1$-AA achieves similar or better results than the worst-case over all competitors.

\section{Conclusion}
We have shown that the proper incorporation of the box constraints in $l_1$-PGD attacks using the correct projection and an adaptive sparsity level motivated by the derived steepest descent direction leads to consistent improvements. Moreover, our $l_1$-APGD is parameter-free and adaptive to the given budget. Using $l_1$-APGD\SPSB{}{CE} in AT yields up to our knowledge the most robust $l_1$-model for $\epsilon=12$. We hope that reliable assessment of $l_1$-robustness via our $l_1$-AutoAttack fosters research for this particularly difficult threat model. An interesting point for future work is to study how and why APGD\SPSB{}{CE} can avoid the failure of $l_1$-adversarial training. 

\section*{Acknowledgements}
The authors acknowledge support from the German Federal Ministry of Education and Research (BMBF) through the Tübingen AI Center (FKZ: 01IS18039A), the DFG Cluster of Excellence ``Machine Learning – New Perspectives for Science'', EXC 2064/1, project number 390727645, and by DFG grant 389792660 as part of TRR 248.


\bibliographystyle{icml2021}

\clearpage

\appendix
\section*{Structure of the appendix}
We provide additional details and experiments which could not be included in the main part. In summary, \begin{itemize} \item Sec.~\ref{sec:app_proofs} contains the proofs omitted above and experiments in support of the effect of the exact projection $P_S(u)$ compared to the approximation $A(u)$ and about the expected sparsity of the steepest descent direction,
\item in Sec.~\ref{sec:app_adv_training} we analyse our observations about catastrophic overfitting in adversarial training wrt $l_1$.
\item Sec.~\ref{sec:app_square} describes the algorithm of our $l_1$-Square Attack,
\item in Sec.~\ref{sec:app_exps} we provide the details of models and attacks used in Sec.~\ref{sec:exp},
\item Sec.~\ref{sec:app_additional_exps} contains experiments 
with a larger threshold, on other datasets, an ablation study on the importance of tuning the parameter of the sparsity $k$ of the updates in SLIDE.

\end{itemize}

\section{Omitted proofs} \label{sec:app_proofs}
In the following we provide the missing proofs from the main paper. For the convenience of the reader we state the propositions and lemmas again.
\begin{customprop}{\ref{pro:exact-projection}} 
The projection problem \eqref{eq:exact-projection} onto $S=B_1(x,\epsilon)\cap H$ can be solved in $O(d \log d)$ with
solution
\[ z^*_i 
= \begin{cases} 1                            & \textrm{ for } u_i \geq x_i \textrm{ and } 0 \leq \lambda_e^* \leq u_i-1\\
                        u_i - \lambda^*_e           & \textrm{ for } u_i \geq x_i \textrm{ and } u_i-1 < \lambda^*_e \leq u_i-x_i\\
                        x_i                         & \textrm{ for } \lambda^*_e > |u_i-x_i|\\
												u_i + \lambda^*_e           &  \textrm{ for } u_i \leq x_i \textrm{ and } -u_i < \lambda^*_e \leq x_i-u_i\\
												0                           &  \textrm{ for } u_i \leq x_i \textrm{ and }  0 \leq \lambda^*_e \leq -u_i\\
												\end{cases},\]
												where $\lambda^*_e \geq 0$. With $\gamma \in \R^d$ defined as
\[ \gamma_i=\max\{-x_i \mathrm{sign}(u_i-x_i),(1-x_i)\mathrm{sign}(u_i-x_i)\},\]							
it holds 	$\lambda_e^*=0$ if $\sum_{i=1}^d \max\{0,\min\{|u_i-x_i|,\gamma_i\}\leq \epsilon$  and otherwise $\lambda^*_e$
is the solution of 
\[ \sum_{i=1}^d \max\big\{0,\min\{|u_i-x_i|-\lambda^*_e,\gamma_i\}\big\}=\epsilon.\]
\end{customprop}
\begin{proof}
By introducing the variable $w'=z-x$ we transform \eqref{eq:exact-projection} into
\[ \hl{\minop_{w' \in \R^d}} \frac{1}{2}\norm{u-x-w'}^2_2  \; \textrm{ s.th. } \; \norm{w'}_1 \leq \epsilon, \quad w'+x \in [0,1]^d.\]
Moreover, by introducing $w=\mathrm{sign}(u-x)w'$ we get 
\begin{align*}
\hl{\minop_{w \in \R^d}} &\;\frac{1}{2}\sum_{i=1}^d (|u_i-x_i|-w_i)^2  \\ \textrm{ s.th. } \; &\sum_{i=1}^d w_i \leq \epsilon, \quad w_i\geq 0, \quad \mathrm{sign}(u_i-x_i)w_i+x_i \in [0,1].
\end{align*} 
The two componentwise constraints can be summarized with 
\[ \gamma_i:=\max\{-x_i \sign(u_i-x_i),(1-x_i)\sign(u_i-x_i)\}\]
as $w_i \in [0,\gamma_i]$.
Note that if $\gamma_i=0$ this fixes the variable $w_i=0$ and we then remove this variable from the optimization problem. Thus wlog
we assume in the following that $\gamma_i>0$.
The corresponding Lagrangian becomes
\begin{align*}
    L(w,\alpha,\beta,\lambda_e)=&\frac{1}{2}\sum_{i=1}^d (|u_i-x_i|-w_i)^2 + \lambda_e(\inner{\ones,w}-\epsilon)\\ &+ \inner{\alpha,w-\gamma}-\inner{\beta,w},
    \end{align*} 
which yields the KKT optimality conditions for the optimal primal variable $w^*$ and dual variables $\alpha^*,\beta^*,\lambda_e^*$
\begin{align*}
 \nabla_w L_i = w^*_i - |u_i-x_i| + \lambda_e^*  +\alpha^*_i -\beta^*_i&=0\\
              \lambda_e^* ( \sum_{i=1}^d w^*_i - \epsilon)&=0\\
							\alpha^*_i (w_i-\gamma_i)  =0, \qquad \beta^*_i w_i              &=0\\
							\lambda_e^*  \geq 0,\; \alpha^*_i \geq 0,\; \beta^*_i & \geq 0
\end{align*}
Thus $\beta^*_i>0$ implies $w^*_i=0$ and with $\gamma_i>0$ this yields $\alpha^*_i=0$ and thus $\beta^*=\lambda_e-|u_i-x_i|$ and we get
\[ \beta^*_i=\max\{0,\lambda_e^*-|u_i-x_i|\}.\]
On the other hand $\alpha^*_i>0$ implies $w^*_i=\gamma_i$ and $\beta^*_i=0$ and thus $\alpha^*_i=|u_i-x_i|-\gamma_i-\lambda_e^*$ and thus
\[ \alpha^*_i=\max\{0,|u_i-x_i|-\gamma_i-\lambda_e^*\}.\]
Thus we have in total
\begin{align*}
w^*_i = \begin{cases} \gamma_i & \textrm{ if } |u_i-x_i|-\gamma_i-\lambda_e^* >0\\ 0 & \textrm{ if } \lambda_e^*-|u_i-x_i|>0 \\ |u_i-x_i|-\lambda_e^* & \textrm{ else}.\end{cases}
\end{align*}
which can be summarized as $w^*_i=\max\{0,\min\{|u_i-x_i|-\lambda_e^*,\gamma_i\}\}$. Finally, if $\sum_{i=1}^d \max\{0,\min\{|u_i-x_i|,\gamma_i\}\}<\epsilon$ then $\lambda_e^*=0$
is optimal, otherwise $\lambda_e^*>0$ and we get the solution from the KKT condition
\begin{equation}\label{eq:opt-equality} \sum_{i=1}^d \max\{0,\min\{|u_i-x_i|-\lambda_e^*,\gamma_i\}\}=\epsilon.
\end{equation}
Noting that 
\[ \phi(\lambda_e)=\max\{0,\min\{|u_i-x_i|-\lambda_e,\gamma_i\}\}\]
is a piecewise linear and monotonically decreasing function in $\lambda_e$, the solution can be found by sorting the union of $|u_i-x_i|-\gamma_i$ and 
$|u_i-x_i|$ in non-decreasing order $\pi$ and then starting with $\lambda_e=0$ and then going through the sorted list until $\phi(\lambda_e)<\epsilon$. In this case one has identified the interval which contains the optimal solution $\lambda_e^*$ and computes the solution with the ``active'' set of components
\[ \{ i \,|\, 0< |u_i-x_i|-\lambda_e <\gamma_i\},\]
via Equation \eqref{eq:opt-equality}. The algorithm
is provided in Algorithm \ref{alg:projl1box}, where the main complexity is the initial sorting step $O(2d \log(2d))$ and some steps in $O(d)$ so that the total complexity is $O(2d \log(2d))$.
Once we transform back to the original variable of \eqref{eq:exact-projection} we get with the form of $\gamma$ the solution
\begin{align*}
z^*_i &= x_i + \sign(u_i-x_i)w^*_i \\
&= \begin{cases} 1                            & \textrm{ for } u_i \geq x_i \textrm{ and } 0 \leq \lambda_e^*
\leq u_i-1\\
                        u_i - \lambda^*_e           & \textrm{ for } u_i \geq x_i \textrm{ and } u_i-1 < \lambda^*_e \leq u_i-x_i\\
                        x_i                         & \textrm{ for } \lambda^*_e > |u_i-x_i|\\
												u_i + \lambda^*_e           &  \textrm{ for } u_i \leq x_i \textrm{ and } -u_i < \lambda^*_e \leq x_i-u_i\\
												0                           &  \textrm{ for } u_i \leq x_i \textrm{ and }  0 \leq \lambda^*_e \leq -u_i\\
												\end{cases}
\end{align*} 
\end{proof}

In order to argue about the approximate projection $A(u)$
we need the same analysis for the $l_1$-projection which can be derived in an analogous way and the solution for the projection onto $B_1(x,\epsilon)$ in \eqref{eq:l1-projection} has the form:
\[ z_1 = \begin{cases} u_i - \lambda^*_1 & \textrm{ for } u_i \geq x_i \textrm{ and } \lambda^*_1 \leq |u_i-x_i| \\
                        x_i                         & \textrm{ for } \lambda^*_1>|u_i-x_i|\\
												u_i + \lambda^*_1 &  \textrm{ for } u_i \leq x_i \textrm{ and } \lambda^*_1 \leq |u_i-x_i| \end{cases},\]
where the optimal $\lambda^*_1\geq 0$ is equal to $0$ if $\sum_{i=1}^d \max\{0,|u_i-x_i|\}\leq \epsilon$ and otherwise it fulfills
\[ \sum_{i=1}^d \max\{0,|u_i-x_i|-\lambda_1^*\}= \epsilon.\]

The two prior versions of PGD  \cite{TraBon2019,maini2020adversarial} for the $l_1$-threat model use instead of the exact projection $P_S$ the approximation $A:\R^d \rightarrow S$
\[ A(u) = (P_H \circ P_{B_1(x,\epsilon)})(u).\]
The following proof first shows that $A(u) \in S$. 
However, it turns out that the approximation $A(u)$ ``hides'' parts of $S$ due to the following property.
Note that the condition is is slightly deviating from the one of Lemma \ref{le:proj-properties} due to a corner case when $\norm{u-x}_1=\epsilon$. Thus we know require $\norm{u-x}_1>\epsilon$ which then implies that $\norm{P_S(u)-x}_1=\epsilon$.
\begin{customlemma}{\ref{le:proj-properties}}
It holds for any $u \in \R^d$,
\[  \norm{P_{S}(u) - x}_1 \geq \norm{A(u)-x}_1.\]
In particular, if $P_{B_1(x,\epsilon)}(u) \notin H$
and $\norm{u-x}_1>\epsilon$ 
and one of the following conditions holds
\begin{itemize}
    \item $\norm{P_S(u)-x}_1=\epsilon$
    \item $\norm{P_S(u)-x}_1<\epsilon$ and $\exists u_i \in [0,1]$ with $u_i\neq x_i$
\end{itemize}
then
\[  \norm{P_S(u) - x}_1 > \norm{A(u)-x}_1.\]
\end{customlemma}
\begin{proof}
It holds $A(u) \in H$ but also $A(u) \in B_1(x,\epsilon)$ as with $z_1 := P_{B_1(x,\epsilon)}(u)$
\[ \norm{z_1-x}\leq \epsilon,\]
and it holds with $P_H(z)=\max\{0,\min\{z,1\}\}$ and $z_A:=P_H(z_1)=A(u)$ that
\[ |z_{1,i} - x_i | = |z_{1,i} - z_{A,i} + z_{A,i} - x_i| = |z_{1,i} - z_{A,i}| + |z_{A,i} - x_i|,\]
which follows as if $z_{1,i}>1$ then $z_{1,i}-z_{A,i}=z_{1,i}-1 > 0$ and $1-x_i\geq 0$ whereas if $z_{1,i}<0$ then
$z_{1,i}-z_{A,i}=z_{1,i}-0 < 0$ and $0-x_i\leq 0$. Thus we get
\begin{equation}\label{eq:l1-equality}
    \norm{z_1-x_1}_1 = \norm{z_1-z_A}_1 + \norm{z_A-x}_1,
\end{equation}
Thus it holds $\norm{z_A-x}_1 \leq \norm{z_1-x}_1 \leq \epsilon$
and we get $A(u) \in H \cap B_1(x,\epsilon)$.
 and thus it is a feasible point of the optimization problem in \eqref{eq:exact-projection} and by the
 optimality of $z_e:=P_{B_1(x,\epsilon) \cap H}(u)$ it follows
 \[ \norm{z_e - u}_2 \leq \norm{A(u)-u}_2.\]

If $\norm{z_1-x}_1<\epsilon$ then $z_1=u$ and thus with
\eqref{eq:l1-equality} one gets $\norm{z_A-x}_1<\epsilon$ as well. In this case $z_A$ is the optimal projection if we just had the box constraints. However, as $\norm{z_A-x}_1<\epsilon$ it is also feasible for \eqref{eq:exact-projection} and thus optimal, $z_e=z_A$.
Moreover, if $\norm{u-x}_1=\epsilon$ then $z_1=u$ and thus
with the same argument we get $z_e=z_A$.
On the other hand if $\norm{z_1-x}_1=\epsilon$ and $z_1 \in H$, then $z_1$ is feasible for \eqref{eq:exact-projection} and the minimum over
a larger set and thus $z_e=z_1=z_A$.

Suppose now that $\norm{u-x}_1>\epsilon$ and $z_1 \notin H$. Then $\norm{z_1-u}_1=\epsilon$ and there exists $\lambda_1^* >0$ (optimal solution of the $l_1$-projection problem) such that
\[ z_A = \begin{cases} \min\{u_i - \lambda^*_1,1\} & \textrm{ for } u_i \geq x_i \textrm{ and }  \lambda^*_1 \leq |u_i-x_i|\\
                        x_i                         & \textrm{ for } \lambda^*_1>|u_i-x_i| \\
											 \max\{u_i + \lambda^*_1,0\} &  \textrm{ for } u_i \leq x_i \textrm{ and } \lambda^*_1 \leq |u_i-x_i| \end{cases},\]
where we have used that $u_i-\lambda^*_1\geq x_i$ for $u_i\geq x_i$ resp. $u_i+\lambda^*_1\leq x_i$ for $u_i\leq x_i$.
Moreover, as $z_1 \notin H$ this implies that $\norm{z_A-x}_1<\norm{z_1-x}_1=\epsilon$. Then if $\norm{z_e-x}_1=\epsilon$ (which implies $\lambda^*_e \leq \lambda^*_1$) there is nothing to prove (we get $\norm{z_e-x}_1 > \norm{z_A-x}_1$ and this represents the first condition in the Lemma for strict inequality) so
suppose that $\lambda^*_e=0$ (that is $\norm{z_e-x}_1<\epsilon$) and thus
\[ z_e = \begin{cases} \min\{ u_i,1\} & \textrm{ for } u_i \geq x_i \textrm{ and } |u_i-x_i|\geq 0\\
                        x_i                         & \textrm{ for } |u_i-x_i|<0\\
												\max\{u_i,0\} &  \textrm{ for } u_i \leq x_i \textrm{ and } |u_i-x_i|\geq 0\end{cases}.\]
Then we get 
\[ |(z_e)_i - x_i| \geq |(z_A)_i-x_i| \quad \forall i=1,\ldots,d.\]
and thus $\norm{z_e-x}_1\geq \norm{z_A-x}_1$. In fact, a stronger property can be derived under an extra condition on $u$. As $\norm{z_1-x}_1=\epsilon$ it must hold $\norm{u-x}_1\geq \epsilon$
and thus $u \neq x$. Now suppose that wlog $u_i > x_i$ and $u_i \in [0,1]$ then  $(z_A)_i = u_i-\lambda^*_1$ if $|u_i-x_i|\geq \lambda^*_1$ or
$(z_A)_i= x_i$ if $|u_i-x_i|<\lambda^*_1$. However, in both cases we get
\[ (z_e)_i - x_i = u_i-x_i > \max\{u_i-x_i-\lambda_1^*,0\}  \geq (z_A)_i-x_i,\]
 and thus $\norm{z_e-x}_1 > \norm{z_A-x}_1$. If $u_i<x_i$ and $u_i \in [0,1]$ then
\[ (z_e)_i-x_i = u_i-x_i <  \min\{u_i-x_i-\lambda^*_1,0\} \leq (z_A)_i-x_i\]
and thus $\norm{z_e-x}_1>\norm{z_A-x}_1$.

\end{proof}\\

\begin{figure}
    \centering
    \includegraphics[width=0.49\textwidth]{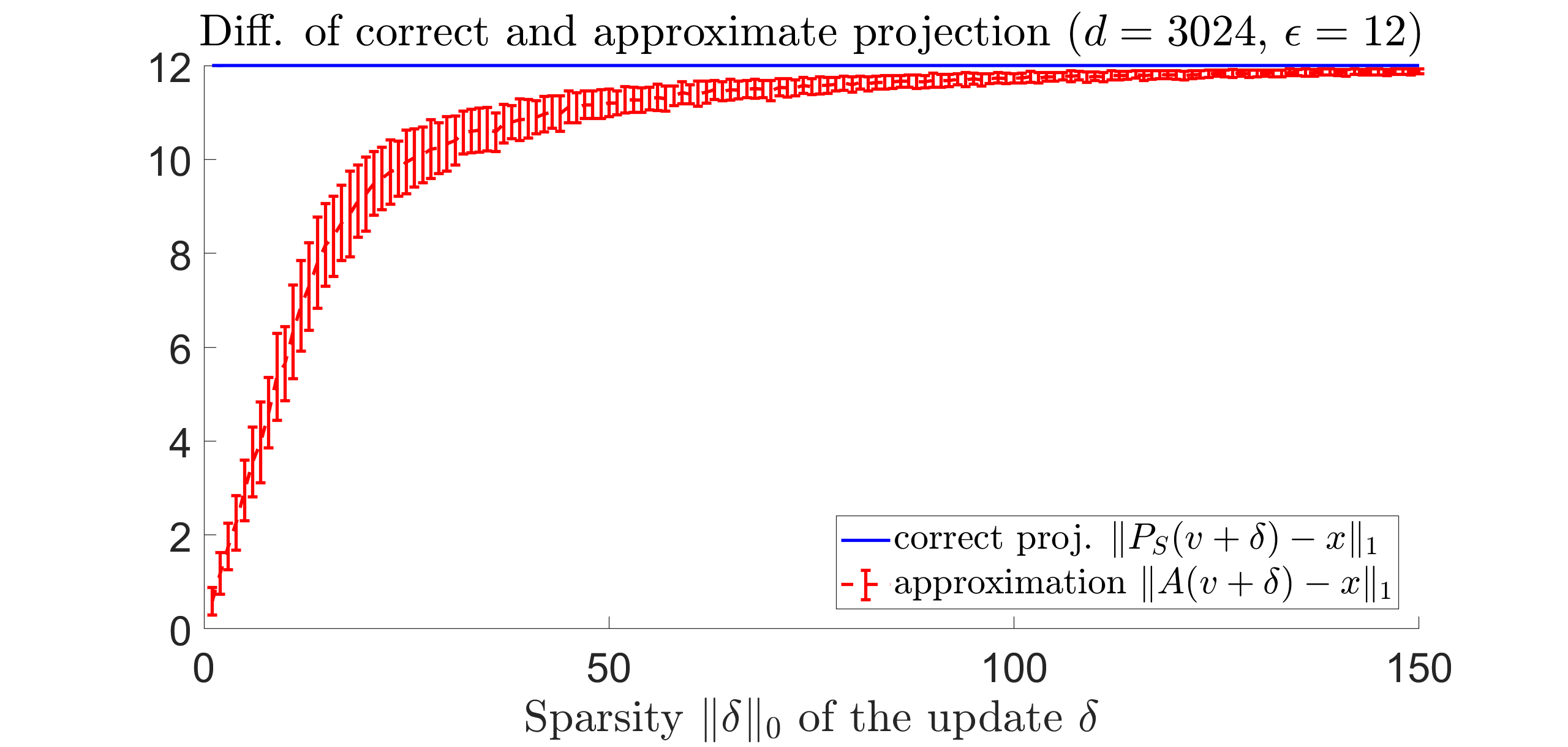}
    \caption{In order to simulate update steps of PGD, we sample target points $x \sim  \mathcal{U}([0,1]^d)$ and $w \in \Nc(0,1)$ and define $v=\P_S(w)$ and project the point $u=v+\epsilon\delta$, where $\delta$ is generated as the descent step in \eqref{eq:sparse_update_step} for different sparsity levels $k$.
    Then we plot $\norm{P_S(u)-x}_1$ and $\norm{A(u)-x}_1$
    for varying sparsity $k$ of the update (for each level $k$ we show average and standard deviation over $100$ samples). It can clearly be seen that the approximate projection is highly biased towards interior regions of the set $S=B_1(x,\epsilon)\cap [0,1]^d$ in particular for small levels of $k$ whereas the correct projection stays on the surface of $S$. Note that SLIDE uses $k=31$ (corresponds to $99\%$ quantile) and thus is negatively affected by this strong bias as effectively large portions of the threat model $S$ cannot be easily explored by SLIDE.}
    \label{fig:ProjRatio}
\end{figure}

In Figure \ref{fig:ProjRatio} we simulate the projections after the steepest descent step \eqref{eq:sparse_update_step} (for varying level of sparsity) to get a realistic picture of the influence of the approximation $A(u)$ versus the exact projection $P_S(u)$. For sparse updates (less than $50$) the approximation behaves quite poorly in comparison to the exact projection in the sense that the true projection is located at the boundary of the $l_1$-ball around the target point $x$ whereas $A(u)$ is located far into the interior of the $l_1$-ball. Note that such sparse updates are used in SLIDE \cite{TraBon2019} and using the exact projection improves SLIDE (see Figure \ref{fig:loss-roberror-PGD}).
Next we derive the form of the steepest descent step.
\begin{customprop}{\ref{prop:sol_lin_with_box}} 
Let $z_i = \max\{(1-x_i)\sign(w_i), -x_i\sign(w_i) \}$, $\pi$ the ordering such that $|w_{\pi_i}| \geq |w_{\pi_j}|$ for $i > j$ and $k$ the smallest integer for which $\sum_{i=1}^k z_{\pi_i} \geq \epsilon$, then the solution of \eqref{eq:lin_opt_withbox} is given by \begin{align}\delta^*_{\pi_i} = 
\begin{cases} z_{\pi_i} \cdot \sign(w_{\pi_i})& \text{for}\; i < k,\\ (\epsilon - \sum_{i=1}^{k-1} z_{\pi_i}) \cdot \sign(w_{\pi_k})&\text{for}\;i=k, \\ 0 & \text{for}\; i > k\end{cases}. \end{align}\end{customprop}
\begin{proof}
We introduce the new variable $\alpha_i:=\sign(w_i)\delta_i$, with the convention $\sign(t) = 0$ if $t=0$. Then we get the equivalent optimization problem:
\begin{align}\label{eq:opt-problem3}
 \max\limits_{\alpha \in \R^d_+}&\; \sum_i |w_i|\alpha_i \\
 \textrm{ s.th. }& \; \sum_{i=1}^d \alpha_i \leq \epsilon, \quad \alpha_i\geq 0,\nonumber\\ &\; -x_i \leq \sign(w_i)\alpha_i \leq 1-x_i, \; i=1,\ldots,d.\nonumber
\end{align}
and thus 
\[ 0\leq\alpha_i \leq \max\{-x_i\mathrm{sign}(w_i),(1-x_i)\mathrm{sign}(w_i)\}.\]
Given the positivity of $\alpha$ and the upper bounds the maximum is attained when ordering $|w_i|$ in decreasing order $\pi$ and setting always $\alpha_{\pi_i}$ to the upper bound until the budget $\sum_{i=1}^d \alpha_i=\epsilon$ is attained. Thus
with $k$ being the smallest integer in $[1,d]$ such that $\sum_{i=1}^k u_i \leq \epsilon$ we get the solution
\[ \alpha_{\pi_i}^* = \begin{cases} z_{\pi_i} & \;\textrm{ for }\; i<k,\\ (\epsilon-\sum_{i=1}^{k-1} z_{\pi_i}) 
& \;\textrm{ for } \;i=k,\\ 0 & \;\textrm{ for }\; i>k\end{cases}.\]
Noting that $\delta_i = \mathrm{sign}(w_i)\alpha_i $ we get
\[ \delta_{\pi_i}^* =\begin{cases} z_{\pi_i} \mathrm{sign}(w_{\pi_i})& \;\textrm{ for }\; i<k,\\ (\epsilon-\sum_{i=1}^{k-1} z_{\pi_i}) \mathrm{sign}(w_{\pi_i})
& \;\textrm{ for } \;i=k,\\ 0 & \;\textrm{ for }\; i>k\end{cases}.\]
\end{proof}

Next we show the expected sparsity of the steepest descent step.

\begin{customprop}{\ref{prop:expected_sparsity}}
Let $w \in \R^d$ with $w_i\neq 0$ for all $i=1,\ldots,d$ and $x \in \U([0,1]^d)$ and let $\delta^*$ be the solution from \ref{prop:sol_lin_with_box}. Then 
it holds for any $\frac{d-1}{2}\geq \epsilon>0$, 
\begin{align*}
\Exp\big[\norm{\delta^*}_0\big] \hspace{-0.6mm}
=&\hspace{-0.6mm}  \lfloor \epsilon+1 \rfloor + \hspace{-1.5mm}\sum_{m=
\lfloor \epsilon \rfloor + 2}^d \sum_{k=0}^{\lfloor \epsilon \rfloor} (-1)^k \frac{(\epsilon-k)^{m-1}}{k! \, (m-1-k)!} \\ &
\geq \frac{\lfloor 3 \epsilon \rfloor + 1}{2}.
\end{align*}
\end{customprop}
\begin{proof}
We first note that the components $z_i$ defined in Proposition \ref{prop:sol_lin_with_box} are independent and have distribution $z_i \sim \U([0,1])$, $i=1,\ldots,d$  as both $1-x_i$ and $x_i$ are uniformly distributed and the $x_i$ are independent. As the $z_i$ are i.i.d. the distribution of $k$ is independent of the ordering of $w$ and thus we can just consider the probability $\sum_{i=1}^k z_i \geq \epsilon$. Note that for any $\epsilon >0$, we have $1 \leq k \leq d$.
Moreover, we note that 
\[ k > m \; \Longleftrightarrow \; \sum_{i=1}^m z_i < \epsilon \; \textrm{ and } \; k=d \Longleftrightarrow \sum_{i=1}^d z_i \leq \epsilon,\]
and as $k$ is an integer valued random variable we have
\[ k \geq m \; \Longleftrightarrow \; k > m-1.\]
Thus as $k$ is a non-negative integer valued random variable, we have
\begin{align*}
 \Exp[k]&=\sum_{m=1}^d \Pr( k \geq m) = \sum_{m=1}^d \Pr\big(k > m-1 \big)\\ &= \sum_{m=1}^d \Pr\Big( \sum_{i=1}^{m-1} z_i < \epsilon \Big) = \sum_{m=1}^d \Pr\Big(\sum_{i=1}^{m-1} z_i \leq \epsilon \Big)
\end{align*}
where in the last step we use that $\sum_{i=1}^{m-1} z_i$ is a continuous random variable and thus we add a set of measure zero. The sum of uniformly distributed random variables on $[0,1]$ has the Irwin-Hall distribution with a cumulative distribution function \cite{irwin1927,hall1927} given by
\[  \Pr\Big(\sum_{i=1}^{m-1} z_i \leq \epsilon \Big) = \frac{1}{(m-1)!} \sum_{k=0}^{\lfloor \epsilon \rfloor}(-1)^k \binom{m-1}{k} (\epsilon-k)^{m-1}.\]
Note that the distribution of $\sum_{i=1}^{m-1} z_i$ is symmetric around the mean value $\frac{m-1}{2}$ and thus the median is also $\frac{m-1}{2}$.
We note that for $m=1$ the first sum is empty and thus $\Pr\Big(\sum_{i=1}^{0} z_i \leq \epsilon \Big)=1$ and in general as 
$0\leq z_1 \leq 1$ it holds for $d-1\geq \epsilon\geq m-1$:
\[ \Pr\Big(\sum_{i=1}^{m-1} z_i \leq \epsilon \Big) = 1 \]
Thus 
\[ \sum_{m=1}^d \Pr\Big(\sum_{i=1}^{m-1} z_i \leq \epsilon \Big) = \lfloor \epsilon+1 \rfloor + \hspace{-3mm}  \sum_{m=\lfloor \epsilon+1 \rfloor+1}^d \Pr\Big(\sum_{i=1}^{m-1} z_i \leq \epsilon \Big).\]
As the median is given by $\frac{m-1}{2}$ we get for $\epsilon \geq \frac{m-1}{2}$  
\begin{align*}
 \Pr\Big(\sum_{i=1}^{m-1} z_i \leq \epsilon \Big) \geq \frac{1}{2}
\end{align*} 
and thus
\begin{align*}
 \sum_{m=1}^d \Pr\Big(\sum_{i=1}^{m-1} z_i \leq \epsilon \Big) \geq  & \lfloor \epsilon+1 \rfloor + \frac{\lfloor 2\epsilon+1\rfloor - \lfloor \epsilon+1 \rfloor}{2}\\ &+ 
 \sum_{m=\lfloor 2 \epsilon+1 \rfloor+1}^d \Pr\Big(\sum_{i=1}^{m-1} z_i \leq \epsilon \Big) \\
\geq & \frac{\lfloor \epsilon+1 \rfloor + \lfloor 2\epsilon + 1\rfloor}{2} 
\geq \frac{\lfloor 3\epsilon \rfloor + 1}{2}
\end{align*}
\end{proof}

\begin{figure}
    \centering
    \includegraphics[width=0.45\textwidth]{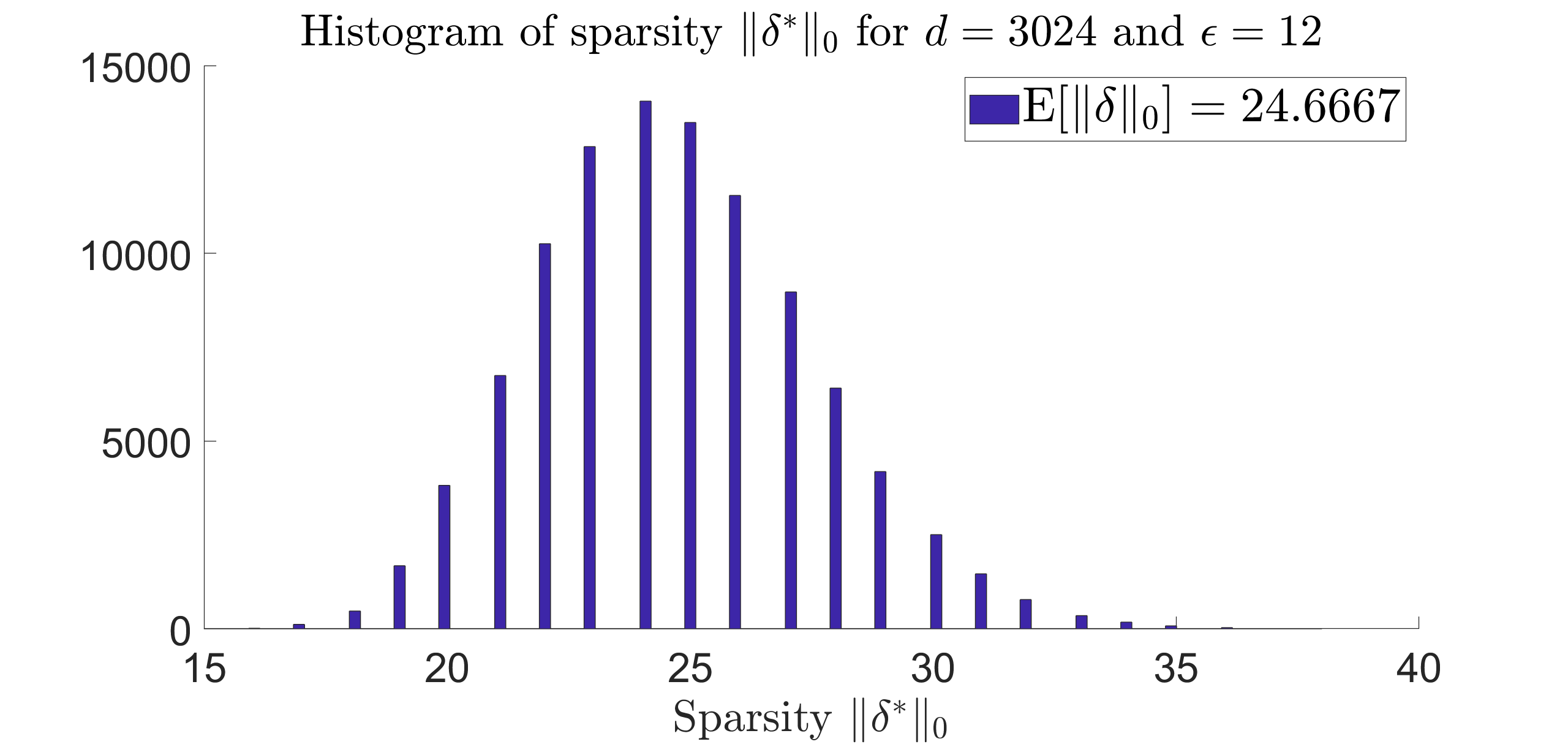}
    \caption{Histogram of the sparsity level $\norm{\delta^*}_0$ of the steepest descent step from Proposition \ref{prop:expected_sparsity} for $d=3024$ and $\epsilon=12$ (histogram computed using $100.000$ samples
    from  $x\in \mathcal{U}([0,1]^d)$ and $w \in \Nc(0,\mathbb{I} 
    )$). The exact expected sparsity can be computed as $\Exp[\norm{\delta^*}_0] \approx 24.6667$.}
    \label{fig:SparsityLevel}
\end{figure}

Note that tighter lower bounds could be derived with more sophisticated technical tools but we decided for the simple argument as we just want to show that the sparsity is non-trivially lower bounded. The exact expression is difficult to evaluate in high dimensions.
In Figure \ref{fig:SparsityLevel} we provide an empirical
evaluation of the sparsity of $\delta^*$ for $d=3024$ and $\epsilon=12$ for $100.000$ samples from the uniform distribution on $[0,1]^d$ ($w$ is drawn from a standard multivariate Gaussian). The exact expected sparsity is about $24.6667$ which is slightly higher than $2\epsilon$. Thus $2\epsilon$ is a reasonable guideline in practice.

\begin{algorithm}[tb]
  \caption{Projection onto $B_1(x,\epsilon)\cap [0,1]^d$}
  \label{alg:projl1box}
\begin{algorithmic}[1]
  \STATE {\bfseries Input:} point to be projected $u$, $x$ and radius $\epsilon$
  \STATE {\bfseries Output:} projection $z$ onto $B_1(x,\epsilon)\cap [0,1]^d$
  
  \STATE $\gamma_i=\max\{-x_i \sign(u_i-x_i),(1-x_i)\sign(u_i-x_i)\}$, $\lambda^*=0$
  \STATE $z_i=\min\{|u_i-x_i|,\gamma_i\}$ 
	\STATE $S=\sum_{i=1}^d z_i$
  \IF{$S > \epsilon$}
	  \STATE sort $t_i=\{ |u_i-x_i|,|u_i-x_i|-\gamma_i\}$ in decreasing order $\pi$ and memorize if it is $|u_i-x_i|$ (category $0$)
		       or $|u_i-x_i|-\gamma$ (category $1$)
		\STATE $M=|\{i \,|\, z_i=|u_i-x_i| \textrm{ and } |u_i-x_i|>0\}|$
		\STATE $\lambda=0$
		\FOR{$j=1$ {\bfseries to} 2d}
		  \STATE $\lambda_{\textrm{old}}=\max\{0,\lambda\}$
			\STATE $\lambda=t_{\pi_j}$
			\STATE $S=S-M (\lambda-\lambda_{\textrm{old}})$
			\IF{category$(\pi_j)=0$}
			  \STATE $M=M+1$
			\ELSE	
			  \STATE $M=M-1$
			\ENDIF
			\IF{$S<\epsilon$}
				\IF{category$(\pi_j)=0$}
			    \STATE $M=M-1$
			  \ELSE	
			    \STATE $M=M+1$
			  \ENDIF
				\STATE $S=S+M (\lambda-\lambda_{\textrm{old}})$
				\STATE $\lambda^*=\lambda_{\textrm{old}} + (S-\epsilon)/M$
			  \STATE \textbf{BREAK}
			\ENDIF
		\ENDFOR
	\ENDIF
  \STATE $z_i=\max\{0,\min\{|u_i-x_i|-\lambda^*,\gamma_i\}\}, \; i=1,\ldots,d$
  \STATE $P_S(u)_i = x_i + \mathrm{sign}(u_i-x_i) \cdot z_i;$ 
\end{algorithmic} \end{algorithm}

\section{Adversarial training wrt $l_1$} \label{sec:app_adv_training}
\begin{figure*}[h!]\centering
\includegraphics[width=1.6\columnwidth]
{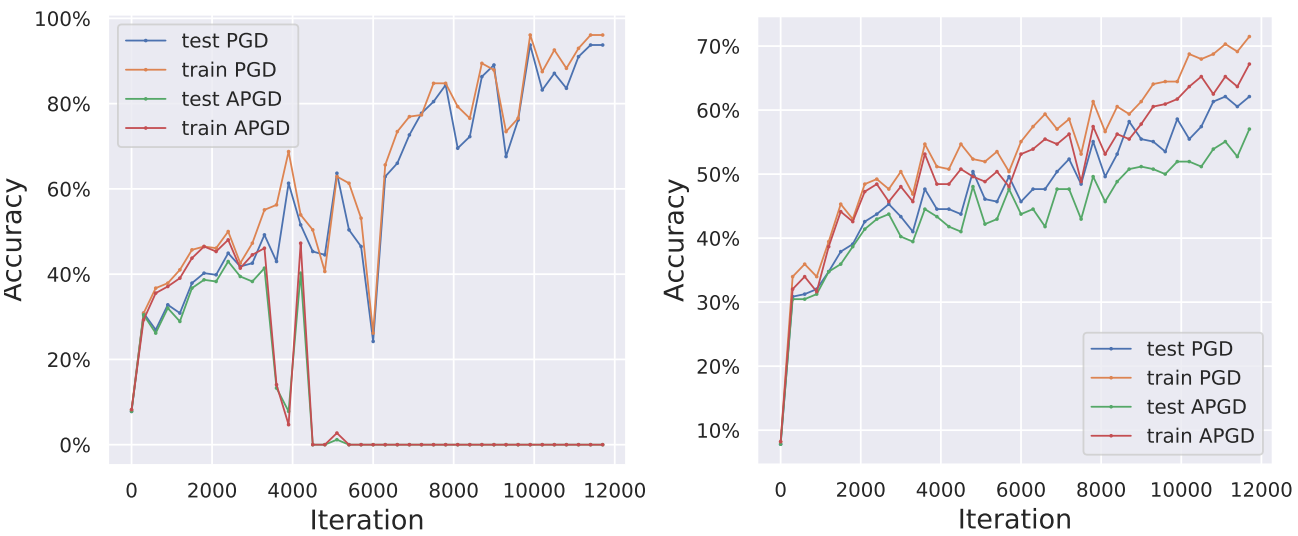}
\caption{\textbf{Left:} Illustration of catastrophic overfitting in $l_1$ adversarial training when using the $10$-step PGD-attack SLIDE \cite{TraBon2019} with $k=0.01$ for training. While the model is robust against the $10$-step SLIDE attack, it is completely non-robust when running our stronger $l_1$-APGD attack with 100 iterations. \textbf{Right:} Catastrophic overfitting does not happen when using $10$-step $l_1$-APGD for adversarial training to train AT-APGD, even when one attacks the model with much stronger attacks at test time, see the evaluation in Table \ref{tab:expensive2}. In both plots we report the robust accuracy on training and test set over the 30 epochs of training.\label{fig:overfitting}}
\end{figure*}

As mentioned in Sec.~\ref{sec:adv_train}, adversarial training \cite{MadEtAl2018} in the $l_1\cap [0, 1]^d$-threat model is more delicate than for $l_\infty$ or $l_2$. In particular, we observe that using the \emph{multi-step} PGD-based attack SLIDE with the standard sparsity of the updates $k=0.01$ (here and in the following $k$ indicates the percentage of non zero elements) leads to catastrophic overfitting when training classifiers on CIFAR-10 with $\epsilon=12$ and 10 steps. Here we mean by catastrophic overfitting that model is robust against the attack used during training even at test time but it fails completely against a stronger attack (APGD in the left part of Figure \ref{fig:overfitting}). This is similar to what has been reported by \cite{Wong2020Fast} in the $l_\infty$-threat model using FGSM \cite{GooEtAl2014}, i.e. a single step method, in adversarial training might produce classifiers resistant to FGSM but not to a stronger PGD attack. Moreover, the robustness against PGD drops abruptly during training, on both training and test sets. In our scenario, as illustrated in Figure~\ref{fig:overfitting} by the left plot, a similar phenomenon happens: when we use SLIDE in adversarial training 
the classifier is initially robust against both SLIDE with 10 steps (blue and yellow curves) and the stronger $l_1$-APGD (green and red) with 100 iterations, but around iteration 4000 the robustness computed with $l_1$-APGD goes close to 0\%, while the model is still very robust against the attack used for training (which is a $10$-step attack and not a single step attack). Conversely, when $l_1$-APGD with 10 steps is used for training (right plot) the model stays robust against both $l_1$-APGD with 100 iterations and SLIDE (and other attacks as shown in Sec.~\ref{sec:exp}).

We could prevent catastrophic overfitting by decreasing the sparsity in SLIDE to $k=0.05$, but this yields poor final robustness (around 50\%) compared to what we achieved using $l_1$-APGD (59.7\% against multiple attacks, see Table~\ref{tab:expensive}), since a weaker attack is used at training time. In fact, we show in Sec.~\ref{sec:app_abl_slide} that values $k> 0.01$ in SLIDE lead to worse performance in most of the cases. We hypothesize that $l_1$-APGD, adapting the sparsity of the updates per point, first prevents the model from seeing only perturbations with similar sparsity and second is able to effectively maximize the loss, allowing adversarial training to perform. We note that even \cite{TraBon2019} observed that using the standard fixed value of $k$ leads to overfitting, and proposed as remedies random sparsity levels and 20 steps. We instead keep the usual 10 steps, and have a scheme which adaptively chooses the sparsity rather than randomly.

\section{$l_1$-Square Attack} \label{sec:app_square}
\begin{algorithm}[tb]
  \caption{Sampling distribution in Square Attack for $l_1\cap [0,1]^d$}
  \label{alg:sampling_square}
\begin{algorithmic}[1]
  \STATE {\bfseries Input:} target point $x$ of shape $h\times h \times c$, radius $\epsilon$, current iterate $\iter{x}{i}$, size of the windows $w$
  \STATE {\bfseries Output:} new update $\delta$
    \STATE $\nu \gets \iter{x}{i} - x$
    \STATE sample uniformly $r_1, s_1, r_2, s_2 \in \{0, \ldots, w - h\}$
	\STATE $W_1:= r_1+1:r_1+h,s_1+1:s_1+h$, $W_2:=r_2+1:r_2+h,s_2+1:s_2+h$ 
	\STATE $\epsilon_{\textrm{unused}} \gets \epsilon - \norm{\nu}_1$,
	\STATE $\eta^* \gets \nicefrac{\eta}{\norm{\eta}_1}$ with $\eta$ as in Eq. (2) of \cite{ACFH2019square}
    \FOR{i = 1 {\bfseries to} c}
        \STATE $\rho \gets  Uniform(\{-1, 1\})$
		\STATE $\nu_{\textrm{temp}} \gets \rho \eta^* + \nicefrac{\nu_{W_1,i}}{\norm{\nu_{W_1,i}}_1}$
		\STATE $\epsilon^i_{\textrm{avail}} \gets \norm{\nu_{W_1 \cup W_2, i}}_1+\nicefrac{\epsilon_{\textrm{unused}}}{c}$
		\STATE $\nu_{W_2,i} \gets 0$, \ \ $\nu_{W_1,i} \gets (\nicefrac{\nu_{\textrm{temp}}}{\norm{\nu_{\textrm{temp}}}_1})\epsilon^i_{\textrm{avail}}$
    \ENDFOR
    \STATE $z \gets P_S(x + 3\nu)$
    \STATE $\delta \gets z - \iter{x}{i}$
\end{algorithmic} \end{algorithm}

\cite{ACFH2019square} introduce the Square Attack, a black-box score-based adversarial attacks, based on random search with square-shaped updates. The key component of such scheme is using an effective distribution to sample at each iteration a new candidate update of the current best point, i.e. achieving the best loss. We adapt the algorithm proposed for $l_2$-bounded attacks and give in Alg.~\ref{alg:sampling_square} the detailed procedure constituting the sampling distribution of our version of Square Attack for the $l_1\cap [0,1]^d$-threat model. If the new point $\iter{x}{i}+\delta$ attains a lower margin loss (since in this case we aim at minimizing the difference between the logit of the correct class and the largest of the others, until a different classification is achieved), then $\iter{x}{i+1} = \iter{x}{i} + \delta$, otherwise $\iter{x}{i+1} = \iter{x}{i}$.
The input $w$ of Alg.~\ref{alg:sampling_square} controls the size of the update and is progressively reduced according to a piecewise constant schedule, in turn regulated by the only free parameter of the method $p$.

\section{Experimental details} \label{sec:app_exps}
We here report details about the experimental setup used in Sec.~\ref{sec:exp}.

\subsection{Models}
\begin{table}[h] \caption{Architecture of the models used in the experimental evaluation on CIFAR-10.} \label{tab:architectures}
\small \centering \vspace{2mm}
\begin{tabular}{L{40mm} | L{30mm}}
\textit{model} & \textit{architecture}\\ \toprule
APGD-AT \textbf{(ours)} & PreAct ResNet-18 \\ \cite{madaan2020learning} & WideResNet-28-10 \\ \cite{maini2020adversarial} - AVG & PreAct ResNet-18 \\ \cite{maini2020adversarial} - MSD & PreAct ResNet-18 \\ \cite{augustin2020} & ResNet-50 \\ \cite{robustness} - $l_2$ & ResNet-50 \\ \cite{rice2020overfitting} & PreAct ResNet-18 \\ \cite{xiao2020enhancing} & DenseNet-121 \\ \cite{kim2020adversarial}\SP{*} & ResNet-18 \\ \cite{CarEtAl19} & WideResNet-28-10 \\ \cite{xu2021adversarial} & ResNet-18 \\ \cite{robustness} - $l_\infty$ & ResNet-50
\\ \bottomrule
\end{tabular} \end{table}

Almost all the models we used are publicly available: the classifiers from \cite{robustness, CarEtAl19, rice2020overfitting, augustin2020} are provided in the library RobustBench \cite{croce2020robustbench}. Those of \cite{maini2020adversarial,xu2021adversarial} can be found in the official pages\footnote{\url{https://github.com/locuslab/robust_union}}\SP{,}\footnote{\url{https://github.com/MTandHJ/amoc}}. Moreover, \cite{madaan2020learning, xiao2020enhancing} made models available via OpenReview\footnote{\url{https://openreview.net/forum?id=tv8n52XbO4p}}\SP{,}\footnote{\url{https://github.com/iclrsubmission/kwta}}. Upon request, \cite{kim2020adversarial} could not give access to the original models out of privacy reasons. Therefore we trained new classifiers using the official code\footnote{\url{https://github.com/Kim-Minseon/RoCL}} following the suggested parameters. For the models denoted in \cite{kim2020adversarial} by ``RoCL'' and ``RoCL+rLE'' we could reproduce both clean and robust accuracy wrt $l_\infty$ with the original evaluation code, while for ``RoCL+AT+SS'' we could match the robust accuracy but not the clean one (here robust accuracy is the one computed using their code). However, we used this last one in our experiments since it is the most robust one wrt $l_1$.

For APGD-AT we trained a PreAct ResNet-18 \cite{he2016identity} with softplus activation function, using cyclic learning rate with maximal value 0.1 for 100 epochs, random cropping and horizontal flipping as training set augmentations. We set 10 steps of APGD for maximizing the robust loss in the standard adversarial training setup \cite{MadEtAl2018}.

Table~\ref{tab:architectures} reports the architecture of every model. As mentioned in Sec.~\ref{sec:exp}, we chose such models to have different architectures, training schemes and even training data, as \cite{CarEtAl19, augustin2020} use unlabeled data in their methods.

\subsection{Attacks}
In the following we report the details of the presented attacks. We use ALMA from Adversarial Library \cite{rony2020adversarial}, with the default parameters for the $l_1$-threat models, in particular $\alpha=0.5$ with 100 iterations, $\alpha=0.9$ with 1000. EAD, B\&B and Pointwise Attack (PA) are available in FoolBox \cite{foolbox}: for EAD we use the $l_1$ decision rule and regularization $\beta=0.01$, for B\&B we keep the default setup, while PA does not have tunable parameters. We reimplemented SLIDE following the original code, according to which we set sparsity of the updates $k=0.01$ for CIFAR-10 and step size $\eta=3.06$, which is obtained rescaling the one used in \cite{TraBon2019} $\eta=2$ for $\epsilon=2000/255$ (see below for a study of the effect of different values of $k$). Also, since no code is available for ZO-ADMM for $l_1$, we adapted the $l_2$ version following \cite{zhao2019design} and then optimized its parameters, using $\rho=2$, $\gamma=0.1$. Finally, we use FAB\SP{T} as available in AutoAttack.

For $l_1$-APGD we fix the values of all parameters to those mentioned in Sec.~\ref{sec:single_eps_apgd} and Sec.~\ref{sec:multi_eps_apgd}. Moreover, we set $p=0.8$ in $l_1$-Square Attack as done in AutoAttack for $l_2$ and $l_\infty$. Thus even $l_1$-AutoAttack can be used without any parameter tuning.

\textbf{Attacks runtime:} Direct comparison of runtime is not necessarily representative of the computational cost of each method since it depends on many factors including implementation and tested classifier. We gave similar budget to (almost all) attacks: for the low budget comparison (see Table~\ref{tab:cheap}) we use 100 iterations, which are equivalent to 100 forward and 100 backward passes for ALMA, SLIDE and $l_1$-APGD, 110 forward and 100 backward passes for B\&B (because of the initial binary search), 150 forward and 100 backward passes for FAB\SP{T}. EAD has instead a 9 times larger budget since we keep the default 9 binary search steps.
As an example, when run using a classifier on CIFAR-10 with PreAct ResNet-18 as architecture, 1000 test points, ALMA and SLIDE take around 25 s, $l_1$-APGD 27 s, FAB\SP{T} 32 s, EAD 105 s, B\&B 149 s.

\section{Additional experiments} \label{sec:app_additional_exps}

\subsection{Effect of sparsity in SLIDE} \label{sec:app_abl_slide}
\begin{table*} \caption{Effect of the sparsity $k$ of the updates in SLIDE \cite{TraBon2019}, whose default value is $k=0.01$.} \label{tab:abl_slide}
\centering \small 
\vspace{2mm}
\begin{tabular}{L{40mm} | *{5}{C{14mm}} | >{\columncolor[rgb]{0.9 0.9 0.9}}C{14mm}}
\textit{model} & $k=0.001$ & $k=0.003$ & $k=0.01$ & $k=0.03$ & $k=0.1$ & APGD\SPSB{}{CE}\\ \toprule APGD-AT \textbf{(ours)} & 
74.5& 70.1& 66.6& \underline{64.4}& 65.7& \textbf{61.3}\\ \cite{madaan2020learning} & 61.9& 58.2& \underline{56.1}& 57.0& 66.2& \textbf{54.7}\\ \cite{maini2020adversarial} - AVG & 71.7& 64.4& 53.8& \underline{51.8}& 64.7& \textbf{50.4}\\ \cite{maini2020adversarial} - MSD & 63.6& 58.5& 53.2& \underline{51.7}& 62.2& \textbf{49.7}\\ \cite{augustin2020} & 60.9& 53.0& \underline{48.8}& 59.3& 74.1& \textbf{37.1}\\ \cite{robustness} - $l_2$ & 49.6& 40.0& \underline{35.1}& 47.4& 67.4& \textbf{30.2}\\ \cite{xiao2020enhancing} & \underline{\textbf{28.2}}& 30.1& 33.3& 36.1& 45.4& 41.4\\ \cite{rice2020overfitting} & 47.0& 37.6& \underline{32.3}& 45.2& 65.2& \textbf{27.1}\\ \cite{kim2020adversarial}\SP{*} & 38.4& 30.6& \underline{25.1}& 34.9& 58.3& \textbf{18.9}\\ \cite{CarEtAl19} & 41.4& 29.9& \underline{19.7}& 24.2& 64.4& \textbf{13.1}\\ \cite{xu2021adversarial} & 35.3& 24.9& \underline{18.2}& 21.1& 58.2& \textbf{10.9}\\ \cite{robustness} - $l_\infty$ & 34.0& 23.6& \underline{14.2}& 17.0& 59.4& \textbf{8.0}
\\ \bottomrule
\end{tabular} \end{table*}
Since the sparsity $k$ of the updates is a key parameter in SLIDE, the PGD-based attack for $l_1$ proposed in \cite{TraBon2019}, we study the effect of varying $k$ on its performance. In Table~\ref{tab:abl_slide} we report the robust accuracy achieved by SLIDE with 5 values of $k\in \{0.001, 0.003, 0.01, 0.03, 0.1\}$ on the CIFAR-10 models used for the experiments in Sec.~\ref{sec:exp}, with a single run of 100 iterations. As a reference, we also show the results of our $l_1$-APGD with the same budget (grey column). We observe that while the default value $k=0.01$ performs best in most of the cases, for 3/12 models the lowest robust accuracy is obtained by $k=0.03$, for 1/12 by $k=0.001$. This means that SLIDE would require to tune the value of $k$ for each classifier to optimize its performance. Moreover, $l_1$-APGD, which conversely automatically adapts the sparsity of the updates, outperforms the best out of the 5 versions of SLIDE in 11 out of 12 cases, with the only exception of the model from \cite{xiao2020enhancing} which is know to present heavy gradient obfuscation and on which the black-box Square Attack achieves the best result (see Sec.~\ref{sec:exp}).

\subsection{Larger threshold $\epsilon$} \label{sec:app_larger_eps}
\begin{table*}
\caption{\label{tab:cheap_larger_eps}\textbf{Low Budget (\boldmath $\epsilon=16$\unboldmath):} Robust accuracy achieved by the SOTA $l_1$-adversarial attacks on 
models for CIFAR-10 
at $\epsilon=16$. 
} \centering \vspace{2mm} {\small
\setlength{\tabcolsep}{1.5pt}
\begin{tabular}{L{40mm} | C{12mm} | *{5}{C{12mm}} >{\columncolor[rgb]{0.9 0.9 0.9}}C{12mm} |C{12mm}>{\columncolor[rgb]{0.9 0.9 0.9}}C{12mm}
}
\textit{model} &clean &EAD & ALMA & SLIDE & B\&B\SPSB{}{} & FAB\SPSB{T}{} & APGD\SB{CE} & PA & Square\\ \toprule APGD-AT \textbf{(ours)} & 87.1& 55.2& 55.3& 59.1& 52.7& 59.7& \textbf{50.6}& 78.5& 66.4\\ \cite{madaan2020learning} & 82.0& 46.2& 50.0& 47.7& 46.0& 48.2& \textbf{45.6}& 70.4& 58.5\\ \cite{maini2020adversarial} - MSD & 82.1& 43.9& 46.2& 44.9& 43.2& 48.5& \textbf{40.9}& 69.8& 57.9\\ \cite{maini2020adversarial} - AVG & 84.6& 43.7& 44.9& 45.9& 43.0& 55.3& \textbf{38.9}& 75.1& 62.9 \\
\bottomrule
\end{tabular}} \end{table*}
We here test that the performance of our attacks at the larger threshold $\epsilon=16$. In Table~\ref{tab:cheap_larger_eps} we run all the methods, with the lower budget, on the four most robust models: one can observe that even with a larger threshold our $l_1$-APGD outperforms the competitors on all models.

\subsection{Other datasets} \label{sec:other_datasets}
\begin{table*} \caption{\label{tab:other_cheap}\textbf{Low Budget:
} Robust accuracy achieved by the SOTA $l_1$ -adversarial attacks on various models for CIFAR-100 and ImageNet in the $l_1$-threat model with radius indicated. The statistics are computed on 1000 points of the test set. PA and Square are black-box attacks. The budget is 100 iterations for white-box attacks ($\times$9 for EAD and +10 for B\&B) and $5000$ queries for our $l_1$-Square-Attack.} \centering \vspace{2mm} {\small
\setlength{\tabcolsep}{1.5pt}
\begin{tabular}{L{40mm} | C{12mm} | *{5}{C{12mm}} >{\columncolor[rgb]{0.9 0.9 0.9}}C{12mm} |C{12mm}>{\columncolor[rgb]{0.9 0.9 0.9}}C{12mm}
}
\textit{model} & clean &EAD & ALMA &SLIDE & B\&B & FAB\SPSB{T}{} & APGD\SPSB{}{CE} & PA & Square\\  \toprule
\multicolumn{10}{l}{}\\
\multicolumn{10}{l}{\textbf{CIFAR-100 (\boldmath $\epsilon=12$\unboldmath)}}\\ \midrule
\cite{rice2020overfitting} - $l_2$ & 58.7& 19.5& 24.4& 17.7& 19.3& 19.6& \textbf{14.6}& 37.0& 23.8\\ \cite{rice2020overfitting} - $l_\infty$ & 54.5& 8.6& 9.9& 6.0& 6.5& 13.0& \textbf{4.5}& 23.0& 10.6
\\ \midrule
\multicolumn{10}{l}{}\\
\multicolumn{10}{l}{\textbf{ImageNet (\boldmath $\epsilon=60$\unboldmath)}}\\ \midrule
\cite{robustness} - $l_2$ & 56.6& 45.6& 50.8& 44.7& 44.4& 44.8& \textbf{43.6}& -& 50.2\\ \cite{robustness} - $l_\infty$ & 61.9& 11.7& 26.6& 11.5& 9.5& 34.6& \textbf{6.3}& -& 23.9
\\ \bottomrule
\end{tabular}} \end{table*}

\begin{table*}\caption{\label{tab:other_expensive}\textbf{High Budget:
} Robust accuracy achieved by the SOTA $l_1$ -adversarial attacks on various models for CIFAR-100 and ImageNet in the $l_1$-threat model with $l_1$-radius indicated. The statistics are computed on 1000 points of the test set. ``WC'' denotes the pointwise worst-case over all restarts/runs of EAD, ALMA, SLIDE, B\&B and, if available, Pointwise Attack. Note that APGD\SPSB{}{CE+T}, the combination of APGD\SPSB{}{CE} and  APGD\SPSB{}{T-DLR} ($5$ restarts each), yields a similar performance as AA (ensemble of APGD\SPSB{}{CE+T}, $l_1$-FAB$^T$ and $l_1$-Square Attack)  with the same or smaller budget than the other individual attacks. 
} \vspace{2mm}
\centering
{\centering \small 
\tabcolsep=1.8pt
\begin{tabular}{L{40mm} | C{13mm} | *{4}{C{13mm}} >{\columncolor[rgb]{0.9 0.9 0.9}}C{13mm}| C{13mm} >{\columncolor[rgb]{0.9 0.9 0.9}}C{13mm}}
\textit{model} & clean &EAD & ALMA & SLIDE & B\&B & APGD\SPSB{}{CE+T} & WC & AA\\ \toprule  
\multicolumn{9}{l}{}\\
\multicolumn{9}{l}{\textbf{CIFAR-100 (\boldmath $\epsilon=12$\unboldmath)}} \\ \midrule
\cite{rice2020overfitting} - $l_2$ & 58.7& 18.4& 17.1& 15.5& 13.1& \textbf{12.1}& 12.7& \textbf{12.1}\\ \cite{rice2020overfitting} - $l_\infty$ & 54.5& 8.1& 5.5& 4.5& 3.0& 3.4& \textbf{2.9}& 3.1
\\ \midrule
\multicolumn{9}{l}{}\\
\multicolumn{9}{l}{\textbf{ImageNet (\boldmath $\epsilon=60$\unboldmath)}}\\ \midrule
\cite{robustness} - $l_2$ & 56.6& 44.1& 45.6& 44.2& \textbf{40.3}& 40.5& \textbf{40.3}& 40.5\\ \cite{robustness} - $l_\infty$ & 61.9& 9.6& 17.1& 8.5& 6.2& 4.6& 5.8& \textbf{4.4}
\\ \bottomrule
\end{tabular}} \end{table*}

We test the effectiveness of our proposed attacks on CIFAR-100 and ImageNet-1k, with $\epsilon=12$ and $\epsilon=60$ respectively, in the same setup of Sec.~\ref{sec:exp}. For CIFAR-100 we use the models (PreAct ResNet-18) from \cite{rice2020overfitting}, for ImageNet those (ResNet-50) of \cite{robustness}: in both cases one classifier is trained for $l_\infty$-robustness, the other one for $l_2$, all are publicly available\footnote{\url{https://github.com/locuslab/robust_overfitting}}\SP{,}\footnote{\url{https://github.com/MadryLab/robustness/tree/master/robustness}}. On ImageNet, because of the different input dimension, we use $k=0.001$ for SLIDE (after tuning it), and we do not run Pointwise Attack since it does not scale. For B\&B we use random images not classified in the target class from the respective test or validation sets as starting points.
We observe that on ImageNet, the gap in runtime between B\&B and the faster attacks increases significantly: for example, to run 100 steps for 1000 test points B\&B takes 3612 s, that is around 14 times more than $l_1$-APGD (254 s). Thus B\&B scales much worse to high-resolution datasets. Also, while B\&B and $l_1$-APGD  have in principle a similar budget in terms of forward/backward passes, one could do much more restarts for $l_1$-APGD in the same time as for B\&B. 

We report in Table~\ref{tab:other_cheap} and Table~\ref{tab:other_expensive} the robust accuracy given by every attack on 1000 points of test set of CIFAR-100 or validation set of ImageNet. Similarly to CIFAR-10, our $l_1$-APGD achieves the best results for all models in the low budget regime (see Table~\ref{tab:other_cheap}) with a significant gap to the second best, either B\&B or SLIDE. Moreover, when using higher computational budget, $l_1$-AutoAttack gives the lowest robust accuracy in 2/4 cases, improving up 1.4\% over WC, the pointwise worst case over all attacks not included in AA, while in the other cases it is only 0.2\% worse than WC, showing that it gives a good estimation of the robustness of the models.

\subsection{Composition of $l_1$-AutoAttack} \label{sec:indiv_attacks}
\begin{table*}[h] \caption{Individual performance of the components of $l_1$-AutoAttack.} \label{tab:indiv_aa}
\centering \small 
\vspace{2mm}
\begin{tabular}{L{40mm} | C{14mm} |*{4}{C{14mm}} | 
C{14mm}
}
\textit{model}& clean &APGD\SPSB{}{CE} & APGD\SPSB{}{T-DLR} & FAB\SPSB{T}{} & Square & AA\\ \toprule 
\multicolumn{7}{l}{}\\
\multicolumn{7}{l}{\textbf{CIFAR-10 (\boldmath $\epsilon=12$\unboldmath)}} \\ \midrule
APGD-AT \textbf{(ours)} & 
87.1& 60.8& 60.8& 65.9& 71.8& \textbf{60.3}\\ \cite{madaan2020learning} & 82.0& 54.2& 52.0& 54.7& 62.8& \textbf{51.9}\\ \cite{maini2020adversarial} - AVG & 84.6& 48.9& 47.5& 59.5& 68.4& \textbf{46.8}\\ \cite{maini2020adversarial} - MSD & 82.1& 48.6& 47.4& 53.5& 63.5& \textbf{46.5}\\ \cite{augustin2020} & 91.1& 34.7& 34.5& 42.4& 56.8& \textbf{31.0}\\ \cite{robustness} - $l_2$ & 91.5& 27.9& 29.3& 32.9& 52.7& \textbf{26.9}\\ \cite{rice2020overfitting} & 89.1& 25.5& 26.3& 30.3& 50.3& \textbf{24.0}\\ \cite{xiao2020enhancing} & 79.4& 32.2& 33.4& 78.6& 20.2& \textbf{16.9}\\ \cite{kim2020adversarial}\SP{*} & 81.9& 17.0& 16.9& 22.2& 36.0& \textbf{15.1}\\ \cite{CarEtAl19} & 90.3& 9.9& 9.9& 21.5& 34.5& \textbf{8.3}\\ \cite{xu2021adversarial} & 83.8& 9.6& 9.3& 17.7& 32.0& \textbf{7.6}\\ \cite{robustness} - $l_\infty$ & 88.7& 6.1& 6.7& 13.0& 28.0& \textbf{4.9}
\\ \midrule
\multicolumn{7}{l}{}\\
\multicolumn{7}{l}{\textbf{CIFAR-100 (\boldmath $\epsilon=12$\unboldmath)}} \\ \midrule
\cite{rice2020overfitting} - $l_2$ & 58.7& 13.4& 13.8& 16.4& 23.8& \textbf{12.1}\\ \cite{rice2020overfitting} - $l_\infty$ & 54.5& 3.8& 4.2& 7.7& 10.6& \textbf{3.1}
\\ \midrule
\multicolumn{7}{l}{}\\
\multicolumn{7}{l}{\textbf{ImageNet (\boldmath $\epsilon=60$\unboldmath)}} \\ \midrule
\cite{robustness} - $l_2$ & 56.6& 42.8& 40.8& 43.1& 50.2& \textbf{40.5}\\ \cite{robustness} - $l_\infty$ & 61.9& 5.7& 5.5& 18.9& 23.9& \textbf{4.4}
\\ \bottomrule
\end{tabular} \end{table*}

We report in Table~\ref{tab:indiv_aa} the individual performance of the 4 methods constituting $l_1$-AutoAttack, recalling that each version of $l_1$-APGD, with either cross-entropy or targeted DLR loss, is used with 5 runs of 100 iterations, FAB\SP{T} exploits 9 restarts of 100 iterations and $l_1$-Square Attack has a budget of 5000 queries. Note that the robust accuracy given by $l_1$-AutoAttack is in all cases lower than that of the best individual attack, which varies across models.

\end{document}